\documentclass[referee]{sn-jnl}

\usepackage{graphicx}%
\usepackage{multirow}%
\usepackage{amsmath,amssymb,amsfonts}%
\usepackage{amsthm}%
\usepackage{mathrsfs}%
\usepackage[title]{appendix}%
\usepackage{xcolor}%
\usepackage{textcomp}%
\usepackage{manyfoot}%
\usepackage{booktabs}%
\usepackage{listings}%
\usepackage{pifont}

\usepackage[square,sort,comma,numbers]{natbib}

\usepackage[nameinlink]{cleveref}

\usepackage{subcaption}

\usepackage{colortbl}
\usepackage{hhline}
\usepackage{makecell}
\usepackage{microtype}
\usepackage{xspace}
\usepackage{nomencl}
\usepackage{nicematrix}
\usepackage{booktabs}
\usepackage{mathrsfs}
\usepackage{tabulary}
\usepackage{pgffor} 
\makenomenclature

\usepackage{etoolbox}
\renewcommand{\nomgroup}[1]{%
  \ifstrequal{#1}{A}{\item[\textbf{Acronyms and Abbreviations}]}{%
  \ifstrequal{#1}{M}{\vspace{5pt}\item[\textbf{Measurements and Noises}]}{%
  \ifstrequal{#1}{V}{\vspace{5pt}\item[\textbf{States and Variables}]}{%
  \ifstrequal{#1}{E}{\vspace{5pt}\item[\textbf{ESKF Matrices}]}{%
  \ifstrequal{#1}{C}{\vspace{5pt}\item[\textbf{Sets and Constants}]}{%
  \ifstrequal{#1}{O}{\item[\textbf{Other Symbols}]}{}}}}}}%
}

\usepackage[lined,commentsnumbered,ruled]{algorithm2e}

\newcommand{\cmark}{\ding{51}}%
\newcommand{\xmark}{\ding{55}}%

\newcolumntype{P}[1]{>{\centering\arraybackslash}p{#1}}
\newcolumntype{M}[1]{>{\centering\arraybackslash}m{#1}}
\def\BibTeX{{\rm B\kern-.05em{\sc i\kern-.025em b}\kern-.08em
    T\kern-.1667em\lower.7ex\hbox{E}\kern-.125emX}}

\newlength{\Oldarrayrulewidth}

\definecolor{mygray}{gray}{0.9}
\newcommand{\grayrow}{\rowcolor[gray]{0.9}}

\definecolor{commentcolor}{RGB}{90, 90, 90}
\definecolor{keywordcolor}{RGB}{0, 64, 128}

\makeatletter
\newcommand{\ie}{\emph{i.e.,}\@ifnextchar.{\!\@gobble}{}}
\newcommand{\eg}{\emph{e.g.,}\@ifnextchar.{\!\@gobble}{}}
\newcommand{\etc}{etc\@ifnextchar.{}{.\@}}
\makeatother

\DeclareMathOperator*{\argmaxA}{arg\,max} 
\DeclareMathOperator*{\argminA}{arg\,min}

\newcommand{\hide}[1]{}

\newcommand{\bdmath}{\begin{dmath}}
\newcommand{\edmath}{\end{dmath}}
\newcommand{\beq}{\begin{equation}}
\newcommand{\eeq}{\end{equation}}
\newcommand{\bdm}{\begin{displaymath}}
\newcommand{\edm}{\end{displaymath}}
\newcommand{\bea}{\begin{eqnarray}}
\newcommand{\eea}{\end{eqnarray}}
\newcommand{\beal}{\beq \begin{array}{ll}}
\newcommand{\eeal}{\end{array} \eeq}
\newcommand{\beas}{\begin{eqnarray*}}
\newcommand{\eeas}{\end{eqnarray*}}
\newcommand{\ba}{\begin{array}}
\newcommand{\ea}{\end{array}}
\newcommand{\bit}{\begin{itemize}}
\newcommand{\eit}{\end{itemize}}
\newcommand{\ben}{\begin{enumerate}}
\newcommand{\een}{\end{enumerate}}

\newcommand{\SO}{\mathrm{SO}}

\newcommand{\Real}{\mathbb{R}}

\newcommand{\SEthree}{\ensuremath{\mathrm{SE}(3)}\xspace}

\newcommand{\RMSE}{\ensuremath{\mathrm{RMSE}}\xspace}

\newcommand{\SOthree}{\ensuremath{\SO(3)}\xspace}

\newcommand{\T}{\mathtt{T}}
\newcommand{\R}{\mathtt{R}}

\newcommand{\residual}{\mathbf{r}}
\newcommand{\trnsp}{\mathsf{T}}

\newcommand{\rotvel}{\boldsymbol\omega}
\newcommand{\acc}{\mathbf{a}}

\newcommand{\tran}{\mathbf{p}}

\newcommand{\vel}{\mathbf{v}}

\newcommand{\bias}{\mathbf{b}}

\newcommand{\gravity}{\mathbf{g}}
\newcommand{\noise}{\boldsymbol\eta}

\newcommand{\F}{\mathbf{F}}
\newcommand{\G}{\mathbf{G}}
\newcommand{\Pcov}{\mathbf{P}}
\newcommand{\Kgain}{\mathbf{K}}

\newcommand{\logmap}{\mathrm{Log}}

\newcommand{\States}{\mathcal{X}}
\newcommand{\covprior}{\mathbf{\Sigma}}
\newcommand{\World}{\text{W}}
\newcommand{\Imu}{\text{B}}
\newcommand{\world}{\text{\tiny{W}}}
\newcommand{\imu}{\text{\tiny{B}}}

\newcommand{\preintRmeas}{\Delta\bar\R}
\newcommand{\preintVmeas}{\Delta\bar\vel}
\newcommand{\preintPmeas}{\Delta\bar\tran}

\newcommand{\toa}{ToA\xspace}
\newcommand{\aoa}{AoA\xspace}
\newcommand{\aod}{AoD\xspace}

\theoremstyle{thmstyleone}%
\theoremstyle{thmstyletwo}%

\theoremstyle{thmstylethree}%

\nomenclature[A]{ESKF}{Error State Kalman Filter}
\nomenclature[A]{ToA}{Time of Arrival}
\nomenclature[A]{AoA}{Angle of Arrival}
\nomenclature[A]{AoD}{Ange of Departure}
\nomenclature[A]{PBCH}{physical broadcast channel}
\nomenclature[A]{SS}{synchronization signals}
\nomenclature[A]{BSs}{Base Stations}
\nomenclature[A]{IMU}{Inertial Measurement Unit}
\nomenclature[A]{PRS}{Positioning Reference Signal}
\nomenclature[A]{PDSC}{Physical Downlink Shared Channel}
\nomenclature[A]{DoF}{Degrees of Freedom}
\nomenclature[A]{RMSE}{Root Mean Square Error}
\nomenclature[A]{EKF}{Extended Kalman Filter}
\nomenclature[A]{CSI}{Channel State Information}
\nomenclature[A]{NN}{Neural Networks}
\nomenclature[A]{DNN}{Deep Neural Network}
\nomenclature[A]{NLOS}{Non-Line-Of-Sight}
\nomenclature[A]{LOS}{Line-Of-Sight}
\nomenclature[A]{MAV}{Micro Aerial Vehicle}
\nomenclature[A]{GNSS}{Global Navigation Satellite Systems}
\nomenclature[Au]{UWB}{Ultra-Wideband}
\nomenclature[A]{NR}{New Radio}
\nomenclature[A]{MAP}{Maximum a Posteriori}
\nomenclature[A]{SLAM}{Simultaneous Localization and Mapping}
\nomenclature[A]{iSAM2}{Incremental Smoothing and Mapping 2}
\nomenclature[A]{ATE}{Absolute Trajectory Error}
\nomenclature[A]{RPE}{Relative Pose Error}
\nomenclature[A]{BW}{Bandwidth}
\nomenclature[A]{RBs}{Resource Blocks}
\nomenclature[A]{SNR}{Signal-to-Noise Ratio}

\nomenclature[A]{OFDM}{Orthogonal Frequency Division Multiplexing}
\nomenclature[A]{RB}{Resource Block}
\nomenclature[A]{PRB}{Resource Block}
\nomenclature[A]{RE}{Resource Element}
\nomenclature[A]{QPSK}{Quadrature Phase Shift Keying}
\nomenclature[A]{PGO}{Pose Graph Optimization}

\nomenclature[V00]{$\States_t$}{State Vector at Time $t$}
\nomenclature[V001]{$\mathscr{X}$}{The Set of All States}
\nomenclature[V01]{$\T$}{Pose Transformation}
\nomenclature[V01]{$\tran$}{Translation Vector}
\nomenclature[V02]{$\R$}{Rotation Matrix}
\nomenclature[V020]{$\mathbf{q}$}{Rotation Quaternion}
\nomenclature[V03]{$\vel$}{Linear Velocity}
\nomenclature[V04]{$\mathbf{b}^g$}{Gyroscope Bias}
\nomenclature[V05]{$\bias^a$}{Accelerometer Bias}
\nomenclature[V06]{$\delta\tran$}{Translation Error}
\nomenclature[V061]{$\imu\rotvel$}{Angular Velocity in IMU Frame}
\nomenclature[V062]{$\world\acc$}{Acceleration in World Frame}
\nomenclature[V07]{$\delta \mathbf{q}$}{Rotation Error Quaternion}
\nomenclature[V08]{$\delta \theta$}{Small Rotation Increments}
\nomenclature[V09]{$\delta\bias^g$}{Gyroscope Bias Error}
\nomenclature[V10]{$\delta\vel$}{Linear Velocity Error}
\nomenclature[V11]{$\delta\bias^a$}{Accelerometer Bias Error}
\nomenclature[V13]{$\hat{\States}_t$}{State Estimate at Time $t$}
\nomenclature[V131]{$\mathbf{d}$}{Real Distance Vector}
\nomenclature[V132]{$\residual^\rho$}{ESKF Innovation}
\nomenclature[V14]{$\preintRmeas_{ij}$}{Preintegrated Measurement of Rotation Between Times $i$ and $j$}
\nomenclature[V15]{$\preintPmeas_{ij}$}{Preintegrated Measurement of Position Between Times $i$ and $j$}
\nomenclature[V16]{$\preintVmeas_{ij}$}{Preintegrated Measurement of Velocity Between Times $i$ and $j$}
\nomenclature[V161]{$\residual_{ij}^\bias$}{Bias Total Residual Between Times $i$ and $j$}
\nomenclature[V162]{$\residual_{ik}^\rho$}{\toa Residual at Time $i$ for Base Station $k$}
\nomenclature[V13]{$\mathcal{Z}^f_t$}{Measurements Used by Factor $f$ at Time $t$}

\nomenclature[V2]{$E_X$}{RMSE Error in x-axis}
\nomenclature[V21]{$E_Y$}{RMSE Error in y-axis}
\nomenclature[V22]{$E_Z$}{RMSE Error in z-axis}

\nomenclature[C00]{$\mathcal{X}$}{Set of State Variables}
\nomenclature[C01]{$\mathcal{T}$}{Set of Time Instances}
\nomenclature[C02]{$\mathcal{F}$}{Set of Factors}
\nomenclature[C03]{$\mathcal{Z}$}{Set of Measurements}
\nomenclature[C04]{$\SOthree$}{Special Orthogonal Group in Three Dimensions}
\nomenclature[C05]{$\SEthree$}{Special Euclidean Group in Three Dimensions}
\nomenclature[C06]{$\World$}{World Frame}
\nomenclature[C07]{$\Imu$}{IMU/Body Frame}
\nomenclature[C08]{$\mathbf{L}_k$}{$k$-th Landmark 3D Position}
\nomenclature[C09]{$\mathrm{N}$}{Total Number of Pose Nodes}
\nomenclature[C10]{$\gravity$}{Gravitational Acceleration}
\nomenclature[C11]{$\rotvel^G$}{Rotational Velocity Due to Earth's Rotation}
\nomenclature[C12]{$c$}{Speed of Light}

\nomenclature[M01]{$\imu\bar\rotvel$}{Angular Velocity Measured by IMU}
\nomenclature[M02]{$\imu\bar\acc$}{Linear Acceleration Measured by IMU}
\nomenclature[M03]{$\bar\rho$}{Time of Arrival Measurements Vector}
\nomenclature[M04]{$\bar{\mathbf{d}}$}{Distance Measurement Vector}
\nomenclature[M05]{$\noise^g$}{Gyroscopic Measurement Noise}
\nomenclature[M06]{$\noise^a$}{Accelerometer Measurement Noise}
\nomenclature[M07]{$\noise^{Dist}$}{Distance Measurement Noise}
\nomenclature[M08]{$\noise^{wg}$}{Random Walk Noise of IMU Gyroscope Bias}
\nomenclature[M09]{$\noise^{wa}$}{Random Walk Noise of IMU Acceleration Bias}
\nomenclature[E]{$\mathbf{H}$}{Measurement Jacobian matrix}
\nomenclature[E]{$\mathbf{R}$}{Covariance matrix of distance measurement noise}
\nomenclature[E]{$\mathbf{K}$}{Kalman gain matrix}
\nomenclature[E]{$\Pcov$}{Covariance matrix of state estimate}
\nomenclature[E]{$\F$}{State transition matrix}
\nomenclature[E]{$\G$}{Input matrix}
\nomenclature[E10]{$\mathbf{Q}^{\text{IMU}}$}{Covariance matrix of IMU noise}

\begin{document}

\title[Article Title]{Graph-Based vs.  Error State Kalman Filter-Based Fusion Of 5G  And Inertial Data For MAV 
Indoor Pose Estimation}


\author*[1]{\fnm{Meisam} \sur{Kabiri}}\email{meisam.kabiri@uni.lu}

\author[1]{\fnm{Claudio} \sur{Cimarelli}}\email{claudio.cimarelli@alumni.uni.lu}

\author[1]{\fnm{Hriday} \sur{Bavle}}\email{hriday.bavle@uni.lu}
\author[1]{\fnm{Jose Luis} \sur{Sanchez-Lopez}}\email{joseluis.sanchezlopez@uni.lu}
\author[1,2]{\fnm{Holger} \sur{Voos}}\email{holger.voos@uni.lu}

\affil*[1]{\orgdiv{Interdisciplinary Center for Security Reliability and Trust (SnT)}, \orgname{University of Luxembourg}, \orgaddress{  \country{Luxembourg}}}

\affil[2]{\orgdiv{Faculty of Science, Technology, and Medicine (FSTM), Department of Engineering}, \orgname{University of Luxembourg}, \orgaddress{ \country{Luxembourg}}}





\abstract{5G New Radio Time of Arrival (\toa) data has the potential to revolutionize indoor localization for micro aerial vehicles (MAVs). However, its performance under varying network setups, especially when combined with IMU data for real-time localization, has not been fully explored so far. In this study, we develop an error state Kalman filter (ESKF) and a pose graph optimization (PGO) approach to address this gap. We systematically evaluate the performance of the derived approaches for real-time MAV localization in realistic scenarios with 5G base stations in Line-Of-Sight (LOS), demonstrating the potential of 5G technologies in this domain. In order to experimentally test and compare our localization approaches, we augment the EuRoC MAV benchmark dataset for visual-inertial odometry with simulated yet highly realistic 5G \toa measurements. Our experimental results comprehensively assess the impact of varying network setups, including varying base station numbers and network configurations, on \toa-based MAV localization performance. The findings show promising results for seamless and robust localization using 5G \toa measurements, achieving an accuracy of 15 cm throughout the entire trajectory within a graph-based framework with five 5G base stations, and an accuracy of up to 34 cm in the case of ESKF-based localization. Additionally, we measure the run time of both algorithms and show that they are both fast enough for real-time implementation.
 }




\maketitle

\printnomenclature
\section{Introduction}
Small drones, so-called Micro Aerial Vehicles (MAVs), are increasingly being used in indoor environments due to their versatility across various applications, such as surveillance tasks or monitoring in intralogistics, to mention only a few. Precise positioning and orientation are paramount for these applications. For instance, MAVs must adeptly navigate through constrained passageways and enclosed spaces in search-and-rescue missions to pinpoint and assist distressed individuals. Similarly, in warehouse scenarios, MAVs must navigate accurately to efficiently carry out tasks like item retrieval and delivery.

While Global Navigation Satellite Systems (GNSS) stand as the predominant positioning technology for outdoor drone applications, their effectiveness diminishes in indoor settings due to the challenges posed by signal attenuation and multipath effects. Consequently, they struggle to provide accurate localization within enclosed spaces. Inertial navigation systems (INSs) present an alternative method for indoor localization, yet they are susceptible to accumulating noise over time, leading to significant deviations in positional accuracy if left uncorrected.

Recent advances in sensor technology have made LiDAR and image-based methods promising alternatives for indoor MAV localization due to their high accuracy and robustness. However, challenges such as computational complexity, the need for sophisticated algorithms, the high cost and still considerable size and weight of LiDAR sensors, and the difficulty of image-based methods in poor-texture or poorly illuminated environments hinder their widespread adoption.

Indoor localization alternatives based on wireless communication, including WLAN, Bluetooth, and Ultra-Wideband (UWB), exhibit limitations in accuracy, scalability, energy efficiency, and cost, as highlighted by various studies~\cite{yang2021survey, zafari2019survey}. For instance, WLAN is notably susceptible to noise interference, Bluetooth faces constraints related to range and precision, and UWB has seen slow progress in standardization. Furthermore, with its focus on low-power communication and constrained range, also Zigbee has limited potential for accurate indoor positioning. These difficulties highlight the need for other MAV indoor positioning technologies to offer sufficient precision and dependability without relying on GNSS signals.

However, recent advances in mobile communications have paved the way for developing location-based services and applications that rely on accurate localization \cite{gupta2023optimal}. In particular, the deployment of fifth-generation cellular networks (5G New Radio, or 5G for short) has opened up new possibilities for indoor localization due to their characteristics, such as high bandwidth, low latency, advanced beamforming techniques, and enhanced coverage~\cite{Kabiri2022, shrivastava2022efficiency, dilli2022hybrid}. Small cell technologies like femtocells and picocells further facilitate comprehensive indoor coverage. Moreover, 5G has been tailored to cater to the diverse requirements of various industries, encompassing fields like industrial automation, asset tracking, and virtual and augmented reality (VR\&AR). The positioning prerequisites outlined by the 3rd Generation Partnership Project (3GPP) span from meter-level accuracy to sub-decimeter precision for vehicle-to-everything (V2X) scenarios (as detailed in~\cite{Dwivedi2021} and related references). In the mm-wave frequency range, 5G benefits from an increased likelihood of Line-Of-Sight (LOS) connections and broader bandwidth, contributing to an elevated positioning accuracy level. 5G employs a dedicated pilot signal termed the Positioning Reference Signal (PRS) for downlink positioning. This signal is utilized to gauge signal delay by cross-correlating the received PRS with a locally generated counterpart at the transmitter. The resulting delay, the so-called Time of Arrival (\toa), is determined by identifying the peak correlation value between the two signals.

\begin{figure}[t]
    \centering
    \includegraphics[width = 1\linewidth]{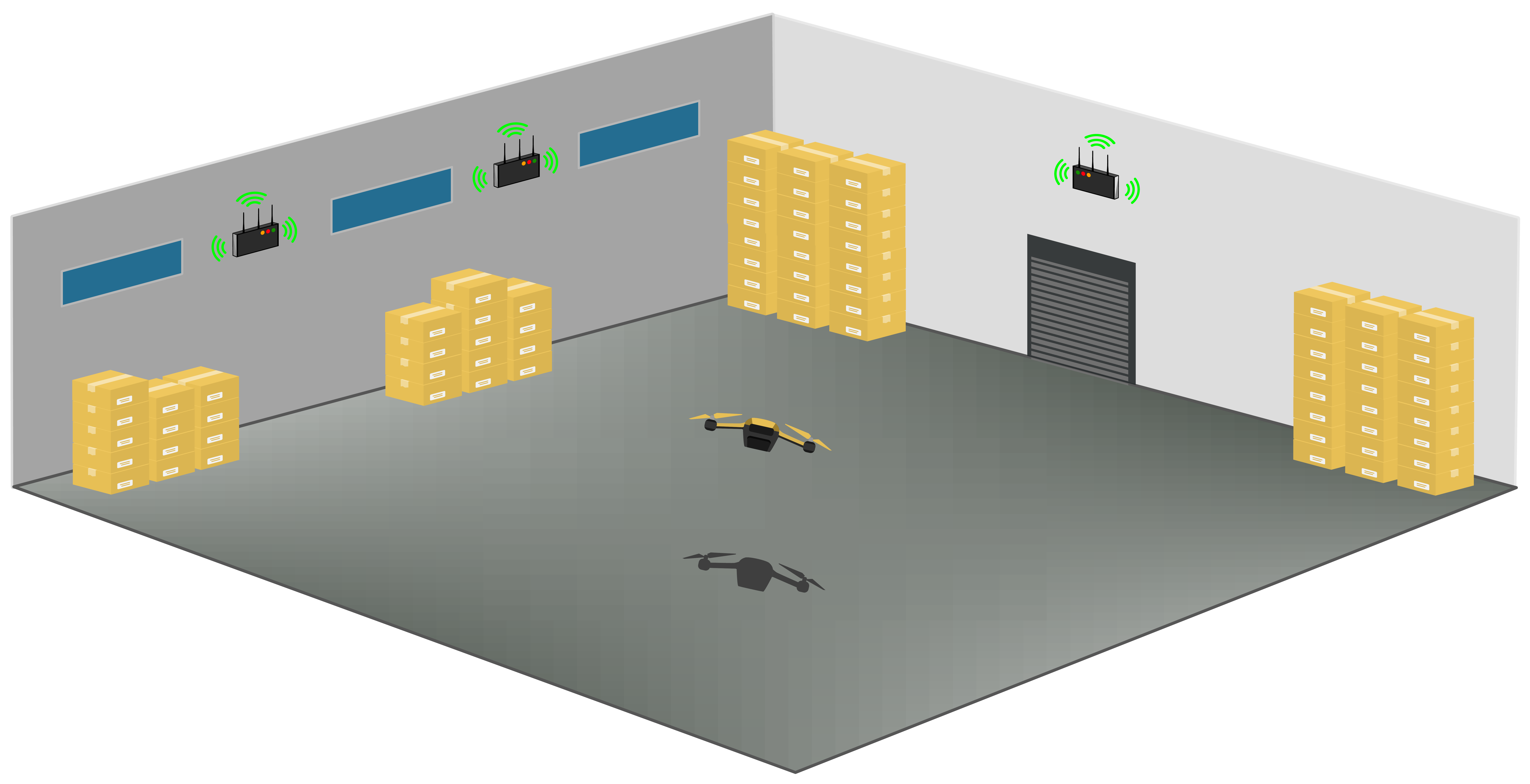}
    \caption{Illustration of a MAV indoor localization example scenario using several 5G base stations.}
    \label{fig:warehouse}
\end{figure}

Therefore, while accomplishing precise and reliable MAV indoor localization continues to be challenging, primarily attributed to indoor environments’ intricate and dynamic nature as well as the highly dynamic motion of the MAV, our work proposes a novel approach to this problem capitalizing on 5G \toa measurements. However, relying solely on 5G \toa data may not yield adequate information for ensuring dependable indoor localization. To enhance reliability and precision, the integration of supplementary sensors becomes imperative. Therefore, we fuse 5G \toa measurements with data obtained from an on-board inertial measurement unit (IMU), providing inertial accelerations and angular velocities of the MAV during flight. By integrating these two types of measurement using advanced optimization techniques, we can achieve accurate and robust real-time pose (i.e., position and orientation) estimations, even when the MAV is maneuvering through complex indoor environments. Fig.~\ref{fig:warehouse} illustrates a typical example scenario in which a MAV has to navigate within a warehouse environment equipped with 5G base stations (BSs).

Thus, our research aims to fuse 5G \toa with IMU measurements, enhancing real-time pose estimation for a flying MAV, specifically focusing on improvements in localization accuracy, scalability and adaptability, real-time performance, and integration with sensor fusion frameworks. This emphasis on drone localization distinguishes our work from studies primarily focused on the 5G side aspects, highlighting our contribution to advancing indoor drone localization techniques. Building upon our prior work ~\cite{kabiri2023pose}, this extended version proposes two novel algorithmic approaches to this sensor fusion problem, namely an Error State Kalman Filter (ESKF) and Pose Graph Optimization (PGO) using a factor graph. Within the ESKF framework, we establish models for the IMU error states and their corresponding covariances. These models facilitate the precise representation of error states as we acquire high-frequency IMU measurements. In the subsequent update stage, we then enhance the estimation process by seamlessly integrating the state estimate and error covariance with lower frequency 5G \toa measurements. This integration mitigates errors and counteracts the intrinsic drift tendencies commonly associated with IMU-based estimations. In the PGO approach, we introduce a novel factor related to the 5G \toa measurements and also apply the idea of IMU preintegration ~\cite{imu_preintegration}. We apply the GTSAM \cite{gtsam} framework to solve the resulting PGO problem.

To validate our solutions, we apply them to the widely used EuRoC MAV benchmark dataset \cite{Burri2016}, which comprises onboard visual-inertial sensor measurements and related ground truth pose data collected during a number of experimental MAV indoor flight sequences. Extending our previous work ~\cite{kabiri2023pose}, we augment six distinct sequences of these flight data (where we use the inertial measurements only and discard the vision data) with related simulated yet highly realistic 5G ToA measurements generated with the help of the QuaDRiGa (quasi-deterministic radio channel generator) channel simulator \cite{jaeckel2017quadriga}. Herein, our simulations include three 5G indoor network configurations with a varying number of base stations (BSs) and assumed Line-Of-Sight (LOS) communication. To ensure precise channel modeling during the simulations, we incorporated the ground truth 6 Degrees-of-Freedom (DoF) pose data from the related EuRoC MAV datasets.

To summarize, the novel contributions provided by this paper are the following:
\begin{itemize}
    \item To the best of our knowledge, this is the first derivation of two novel approaches where 5G \toa and IMU measurements are fused for real-time pose estimation of a MAV during indoor flights: (1) an ESKF-based approach and (2) an approach based on PGO using factor graphs.
    \item We augment a part of the well-known EuRoC MAV benchmark dataset with related simulated 5G \toa measurements. In order to obtain highly realistic simulated data, we use the ground truth pose data of six selected EuRoC MAV flight sequences in a simulation of a 5G indoor network with different configurations and number of BSs, and a very detailed communication channel model provided by the QuaDRiGa simulator.
    \item We present a thorough assessment and comparison of the two novel proposed approaches for MAV indoor pose estimation using the augmented EuRoC MAV benchmark dataset. Herein, we also systematically compare various 5G networks and communication settings across indoor environments. For the assessment, we utilize the two most popular metrics in Simultaneous Localization and Mapping (SLAM), i.e., the Absolute Trajectory Error (ATE) and the Relative Pose Error (RPE) \cite{prokhorov2019measuring}.
    \item We finally show that both proposed approaches for MAV indoor pose estimation achieve high accuracy while the PGO-based approach considerably outperforms the ESKF-based approach in terms of accuracy.
\end{itemize}
 
\section{Related Works}
\label{sec:relatedworks}

The literature on localization using 5G is relatively limited, especially when considering a dynamic target, i.e., the MAV in our case, and the integration of sensor fusion frameworks. The existing literature often relies on simplistic methodologies and scenarios. Ferre et al \cite{ferre2019positioning} compared localization accuracy for different combinations of the 5G network configurations (center frequency, sub-carrier spacing, and PRS comb size) in terms of the Root Mean Square Error (RMSE). Also, their study considered a stationary target and employed multilateration based on PRS-derived \toa data from multiple BSs. 
A study by del Peral-Rosado et al \cite{PeralRosado2016} explored the impact of positioning performance using a 5G network when BSs are linearly placed along a straight roadside. They utilized Gauss-Newton optimization and simulated a vehicle traveling at a constant velocity of 100 km/h on a highway. The study revealed an accuracy of less than 20-25 cm for a communication bandwidth of 50-100 MHz. Additionally, the researchers calculated the \toa by determining the first correlation peak between the PRS and the received signal. Saleh et al \cite{Saleh2022} proposed a time-based position estimation by combining vehicle velocity information and 5G measurements. They evaluated their approach in a simulated urban canyon using Siradel's S\_5GChannel simulator \cite{Siradel_S_5GChannel} and employed an Extended Kalman Filter (EKF) with a constant velocity model for sensor fusion. The study also analyzed the impact of the 5G geometrical setup on EKF position estimation. Another EKF-based positioning framework is proposed by Menta et al \cite{menta2019performance}. The authors leveraged the 5G Angle of Arrival (\aoa) extracted from the communication signal of BSs equipped with multi-array antennas. By utilizing this information, they achieve sub-meter accuracy in localization.
Sun et al \cite{Sun2020} studied localization by combining \aoa estimates from 5G BSs with \toa measurements from GNSS satellites. The authors utilized the Taylor series to linearize the mathematical model. As post-processing, they applied a moving averaging to the raw position estimates to minimize errors. Klus et al \cite{klus2021neural} explored the fusion of beamformed RSS information with GNSS data using Neural Networks (NN), achieving meter accuracy. In a study by Talvitie
et al \cite{talvitie2018positioning}, a simulation-based system was proposed for positioning high-speed trains using 5G technology. The system utilized measurements from both the Angle of Departure (\aod) and \toa from beamformed 5G synchronization signals. Using an EKF to track the train's position, the study found that \toa-based positioning was more accurate than \aod-based. However, combining both measurements improved accuracy, i.e., submeter accuracy could be achieved for more than 75\% of the tracking time.

There are also works whose primary focus is on the communication side, emphasizing the analysis of signal characteristics, like channel estimation, or computing \toa and \aoa using MUSIC (Multiple Signal Classification) algorithms, and then primarily considering a simplified multilateration scenario for localization. 
For instance, Pan et al \cite{pan2022efficient} estimated the \aoa and \toa based on the channel state information (CSI) for indoor positioning. Then, using multilateration, they investigated the position accuracy. Shamaei and Kassas \cite{shamaei2021receiver} proposed an opportunistic \toa estimation approach that exploits 5G synchronization signals (SS) and the physical broadcast channel (PBCH), with broadcast in the downlink channel and without requiring any communication with the network. This block is transmitted periodically and is always on, even when the user is not connected to the network. The authors developed a software-defined receiver (SDR) to extract ranging information from these signals and derived the statistics of the code phase error in a multipath-free environment and in the presence of multipath. They also conducted experiments to evaluate the ranging performance of the proposed SDR with real 5G signals, achieving a ranging error standard deviation of 1.19 m. In a fingerprinting-based method, Zhang et al \cite{zhang2021aoa} utilized \toa, \aoa, and multi-path effects to improve positioning accuracy. They created a fingerprint dataset using 5G \aoa and amplitude information for all paths obtained from channel state data. This dataset was used to train a deep neural network (DNN) as a position estimator. The authors reported an accuracy of approximately one meter, even in non-line-of-sight (NLOS) environments.  
In a more advanced theoretical approach, Chu et al \cite{chu2021vehicle}, Mendrzik et al \cite{mendrzik2018joint} investigated both localization and radio mapping, where the multi-path information is used to estimate the positions of the vehicle as well as the reflectors (obstacles) in the environment simultaneously using factor graphs. Their work, however, lacks network simulations and fusion schemes. 

An Error State Kalman Filter (ESKF) is typically used for highly nonlinear systems, and it can improve the accuracy and robustness of state estimation \cite{9206131,panich2010indirect, SANCHEZLOPEZ201716003}.
Yin et al \cite{Yin2023} used the ESKF and Rauch–Tung–Striebel (RTS) smoother to fuse GNSS and IMU data for localization and achieved good results even in challenging environments. Similarly, Marković et al \cite{9836124} proposed an ESKF-based multi-sensor fusion algorithm for UAV localization in indoor environments, which was able to accurately track the UAV's position using measurements from IMU, LiDAR, visual odometry, and UWB sensors. Mourikis and Roumeliotis \cite{4209642} introduced the Multi-State Constraint Kalman Filter (MSCKF) as an algorithm for estimating the state of a vision-aided inertial navigation (VINS) system. The MSCKF is based on the ESKF and fuses visual and inertial measurements to estimate the vehicle's position, orientation, and velocity. Its performance has been evaluated using several datasets, including the EuRoC Mav dataset \cite{Burri2016}, and the results show that it outperforms previous VINS algorithms in terms of accuracy and robustness.

While filtering is a sequential approach to pose estimation, Pose Graph Optimization (PGO) is a batch-based approach that considers more information over time, leading to more accurate and robust pose estimates \cite{factor_graphs_for_robot_perception}. For example, Mascaro et al \cite{8460193} proposed a PGO-based multi-sensor fusion approach for drone localization. It fuses visual-inertial odometry poses and GPS measurement to infer the 6 DoF pose of the robot in real-time. ORB-SLAM 2 \cite{mur2017orb} and ORB-SLAM 3 \cite{campos2021orb} are popular examples of PGO-based mobile robot localization systems that fuse images and IMU data. ORB-SLAM 2 is a real-time visual SLAM system that uses PGO to estimate the pose of a camera as it moves through an environment. ORB-SLAM 3 is a recent version of ORB-SLAM that includes several improvements, including a more efficient PGO implementation.

Unlike previous approaches that consider localization based on 5G data, we address indoor localization using a factor graph to model the relation among non-homogeneous sensor measurements. We also leverage the advanced IMU preintegration factor \cite{imu_preintegration} to propagate the MAV's 6 DoF pose between two lower frequency \toa measurements, obtaining 6 DoF pose estimates at a higher frequency. Furthermore, to establish a comprehensive benchmark for our graph-based method, we have also implemented an ESKF in this paper. This incorporation enables a direct and insightful comparison between the graph-based approach and the sequential ESKF technique, providing a well-rounded evaluation and comparison of the efficacy and performance of both localization methodologies.

To identify the key research gaps within the literature, we can summarize them as follows:

\par

\begin{itemize}

\item \bf Limited scope:

\begin{itemize}
\item Most existing methods focus on static targets or vehicles with constant velocity, while MAVs are dynamic with complex motion patterns.
\item Existing studies often consider simplistic scenarios lacking real-world complexity.
\item There's a lack of research on sensor fusion frameworks specifically designed for 5G-based MAV localization.
\end{itemize}

\item  \bf Methodological limitations:

\begin{itemize}
\item Many studies rely on basic multilateration using signals from multiple base stations, neglecting advanced algorithms and sensor fusion techniques.
\item  Some approaches focus primarily on communication aspects, analyzing signals and estimating angles/times of arrival without fully addressing localization challenges.
\item  Existing works often use simplified mathematical models that don't accurately capture the complexities of real-world environments and MAV dynamics
\end{itemize}

\item  \bf Accuracy and robustness limitations:

\begin{itemize}
\item Many methods achieve sub-meter accuracy, which might not be sufficient for precise indoor MAV operations.
\item Limited research exists on methods that address the drift inherent in inertial measurement units (IMUs), crucial for MAV localization.
\end{itemize}

\item  \bf Lack of benchmarking and comparison:

\begin{itemize}
\item Existing studies often use different datasets, metrics, and network configurations, making comparisons challenging.
\item Few works directly compare different localization approaches, leaving unclear which methods perform best in specific scenarios.
\end{itemize}

\end{itemize}

Table \ref{tab:research_gaps} provides a concise overview of existing research on 5G-based localization for MAVs, elucidating the principal gaps and limitations in prior approaches. Each study is outlined with details including the primary gap identified, utilization of 5G data, experimental/simulation approach, integration of sensor fusion, localization technique employed, and consideration of fixed or dynamic MAV targets.

\begin{table*}
  \centering

  \caption{Comparison of Related Works in 5G-based MAV Localization}
  \label{tab:research_gaps}
  \setlength{\extrarowheight}{0.5em} 
  \resizebox{1\linewidth}{!}{
  \begin{tabular}{c c c c c c c}

\Xhline{3\arrayrulewidth}
\bf References & \bf Primary gap & \bf 5G data & \bf Simulation/Experiment & \bf Sensor Fusion & \bf Technique & \bf Dynamic Target   \\
\Xhline{1\arrayrulewidth}
\rowcolor{mygray} \cite{ferre2019positioning} & Oversimplified scenario with fixed target & \cmark & Simulation (Qaudriga) &  \xmark &  Lateration &  \xmark \\

                 \cite{PeralRosado2016} &  Simplicity of Approach, with constant velocity assumption  &  \cmark &  Simulation (Qaudriga) &  \cmark & Least square &  \cmark \\

\rowcolor{mygray} \cite{Saleh2022} & Lacks robustness and complexity for real-world MAV scenarios & \cmark & Simulation (S-5GChannel) &  \cmark &  EKF &  \cmark \\

                    \cite{menta2019performance} &  Lacks sensor fusion, limiting accuracy and robustness &  \cmark &  Experiment &  \cmark &  EKF &  \xmark \\
                    
\rowcolor{mygray} \cite{Sun2020} & Simplicity of Approach, Lacks robustness and complexity & \cmark & Simulation (Qaudriga) &  \cmark &   Least Square &  \cmark \\ 

                \cite{klus2021neural} &  Achieves low accuracy (meter-level),  insufficient for many MAV applications &  \cmark &  Simulation (ray tracing) &  \cmark &  NN &  \cmark \\

\rowcolor{mygray} \cite{talvitie2018positioning} & Simplicity of Approach, excluding real MAV data & \cmark & Simulation &  \cmark &  EKF &  \cmark \\ 

               \cite{pan2022efficient} &  Focuses on communication aspects, neglecting sensor fusion and real-world complexities &  \cmark &  Experiment &  \xmark &  Triangulation &  \xmark \\

\rowcolor{mygray} \cite{shamaei2021receiver} & Primarily communication-centric, lacking comprehensive sensor fusion for robust MAV localization & \cmark & Experiment &  \xmark &  Lateration &  \cmark \\ 

               \cite{zhang2021aoa} &  Limited to communication data without sensor fusion &  \cmark &  Simulation &  \xmark &  DNN &  \xmark \\

\rowcolor{mygray} \cite{mendrzik2018joint,chu2021vehicle} & Exclude sensor fusion and real MAV data & \cmark & Not network simulation, theoretical analysis &  \xmark &  PGO &  \cmark \\ 

                \cite{Yin2023} &  No 5G data, not relevant to the scope of 5G-based MAV localization &  \xmark &  Simulation (non-5G data)  &  \cmark &  ESKF &  \cmark \\

\rowcolor{mygray} \cite{9836124} & No 5G data, not relevant to the scope of 5G-based MAV localization& \xmark & Experiment (non-5G data) &  \cmark &  ESKF &  \cmark \\ 

                \cite{4209642} &  No 5G data, not relevant to the scope of 5G-based MAV localization &  \xmark &  Experiment (non-5G data) &  \cmark &  ESKF &  \cmark \\

\rowcolor{mygray} \cite{8460193} & No 5G data, not relevant to the scope of 5G-based MAV localization & \xmark & Simulation (non-5G data) &  \cmark &  PGO &  \cmark \\ 

\cite{mur2017orb, campos2021orb} & No 5G data, not relevant to the scope of 5G-based MAV localization & \xmark & Dataset-based evaluation (non-5G data) &  \cmark &  PGO &  \cmark \\

\Xhline{3\arrayrulewidth}

\end{tabular}
}
\end{table*}

\section{Materials and Methods}
\subsection{5G Signal Structure and PRS Fundamentals for Localization}

This section introduces some preliminary concepts related to 5G that are essential for understanding the \toa configurations and other related topics discussed in Section~\ref{sec:evaluation}.

5G employs Orthogonal Frequency Division Multiplexing (OFDM) with the inclusion of a cyclic prefix, which is a guard band added to the beginning of each OFDM symbol to reduce inter-symbol interference. In the time domain, OFDM breaks down a high-speed stream of digital bits into multiple parallel slower-moving streams. In the frequency domain, OFDM divides a wideband signal into multiple narrowband subcarriers, each of which carries a small portion of the data. Subcarrier spacing, referring to the separation or interval between adjacent subcarriers within the frequency domain, is a crucial parameter in 5G Localization accuracy. 5G offers flexible options for subcarrier spacing, ranging from 15 kHz to 240 kHz, influencing the number of subcarriers within each narrowband subcarrier.

\begin{figure*}
    \centering
    \includegraphics[width = 0.75\linewidth]{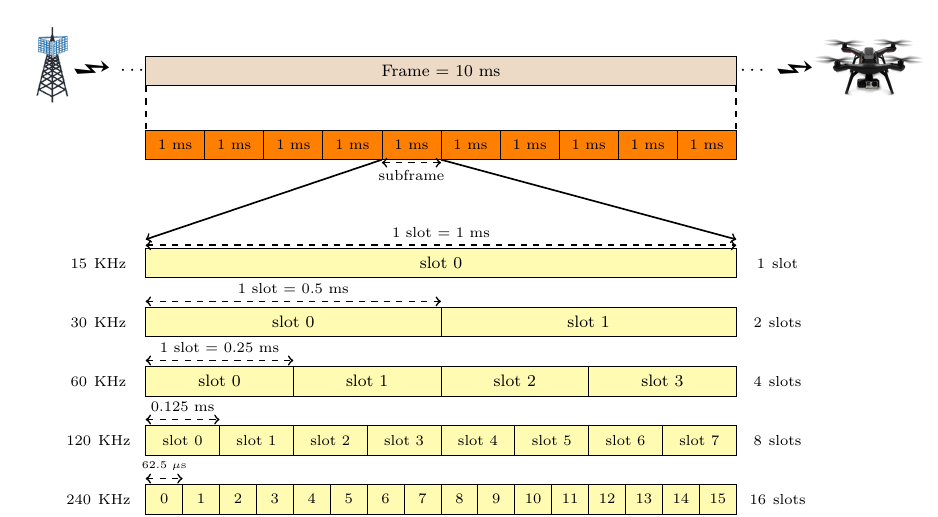}
    \caption{{5G Frame Structure}}
    \label{fig:5gframe}
\end{figure*}

In 5G's downlink transmission, a frame has a duration of 10 milliseconds, which is further divided into ten subframes, each lasting 1 millisecond. Fig.~\ref{fig:5gframe} shows the time-domain structure of the 5G frame, illustrating its reliance on the subcarrier spacing. In the time domain, the choice of subcarrier spacing determines the number of slots in a subframe, each slot contains 14 OFDM symbols. As illustrated in this figure, subframes are subdivided into 1, 2, 4, 8, or 16 slots, depending on the chosen subcarrier spacing, all utilizing a standard cyclic prefix.

\begin{figure}
    \centering
    \includegraphics[width = 0.5\linewidth]{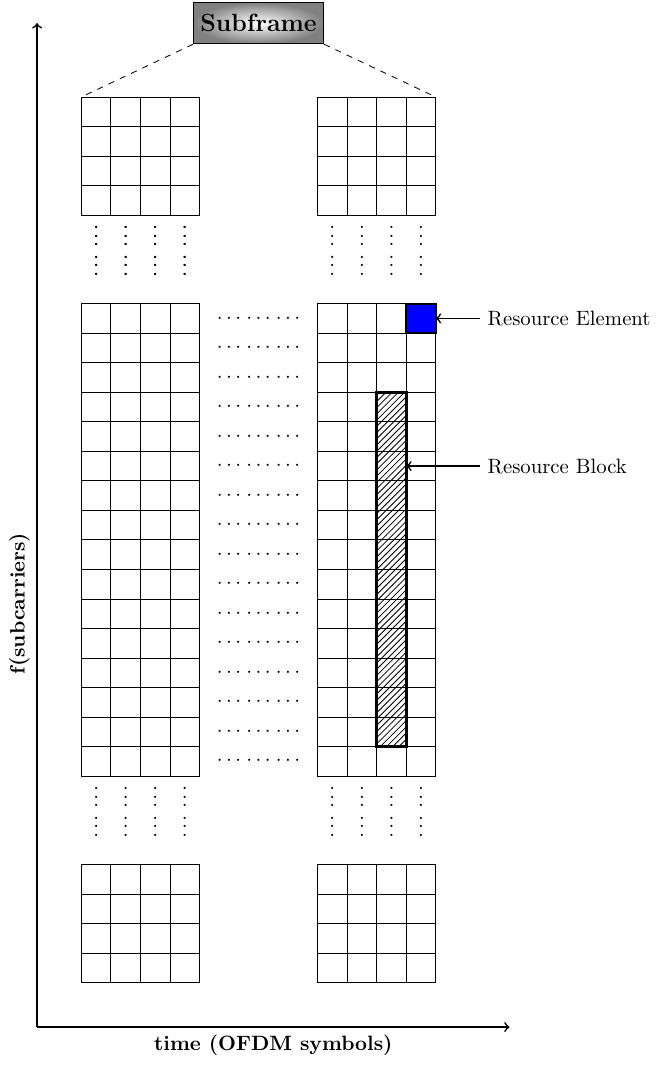}
    \caption{Visualization of the 5G Resource Grid Structure, highlighting Resource Blocks (RBs) and Resource Elements (REs), showcasing the allocation of resources in Resource Grid.}
    \label{fig:rg}
\end{figure}

The resource grid is a two-dimensional structure where time is represented along one axis, and frequency is represented along the other (matrix of subcarriers and OFDM symbols). A resource grid is characterized by one subframe in the time domain and full carrier bandwidth in the frequency domain. This grid is used to allocate resources for communication between the base station (gNodeB) and user devices (User Equipment UE). Resource grids are further divided into resource blocks (RBs). Each RB is composed of a group of subcarriers. The number of subcarriers in an RB depends on the subcarrier spacing. For example, with 15 kHz subcarrier spacing, an RB contains 12 subcarriers. A Physical resource block (PRB) is the smallest unit of resource that can be allocated to a user in 5G. A PRB is defined as a group of consecutive subcarriers in the frequency domain and a group of consecutive OFDM symbols in the time domain. Finally, a resource element (RE) is the smallest unit of resource allocation in the 5G resource grid. Each RE contains one subcarrier and one OFDM symbol. Fig.~\ref{fig:rg} shows the resource grid, PRBs, Rbs, RE, as well as the distribution of the reference signal PRS which will be explained in more detail in the next section. 

For precise positioning 5G supports Positioning Reference Signals (PRS) in downlink. PRS are generated using Gold sequences, which are a type of pseudo-random code, enabling easy reproduction of the PRS at the receiver and resistance to interference. Gold sequences also have good cross-correlation properties, meaning that they have a low correlation with all other sequences in the same set of Gold sequences. PRS symbols then are generated using Quadrature Phase Shift Keying (QPSK) modulation and distributed across time and frequency in the resource grid. The PRS  pattern can be configured with parameters like starting resource element, periodicity, and density. The configuration depends on the specific requirements of the positioning application. One critical configuration that significantly impacts the \toa estimation accuracy is the PRS bandwidth. A wider PRS bandwidth will result in more accurate and robust localization but at the cost of increased interference.

\begin{figure}
    \centering
    \includegraphics[width = 0.5\linewidth]{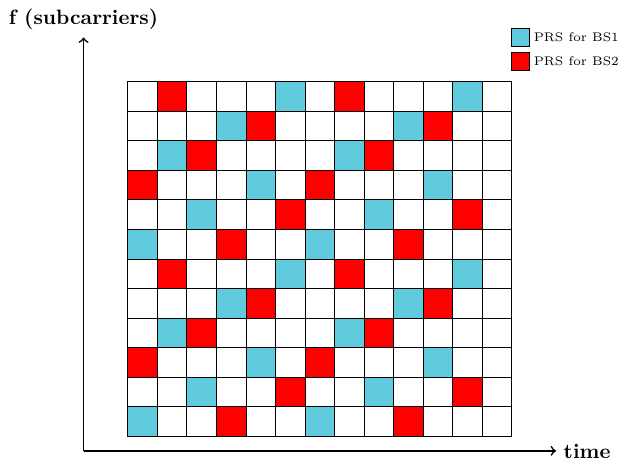}
    \caption{PRS distribution in a physical resource block in 5G NR with two BSs using a comb-6 structure}
    \label{fig:prs}
\end{figure}

Downlink PRS utilizes a comb-like frequency structure by transmitting PRS within a subset of the sub-bands derived from the divided PRS bandwidth. The comb size N in 5G PRS dictates that every Nth subcarrier is used by PRS symbols, with configurable values of N (2, 4, 6, or 12). This effectively diminishes interference from neighboring base stations. It also helps to improve the correlation properties of the PRS signal, which leads to enhanced accuracy and robustness of \toa estimation. PRS mutation is another feature in 5G PRS which can also decrease the interference of PRS from neighboring BSs. Fig.~\ref{fig:prs} illustrates a Physical Resource Block (PRB) featuring PRS symbols configured with a comb-6 structure for two BSs. For more detailed information, we refer to \cite{Dwivedi2021}. Finally, before transmission, the PRS and data signals are OFDM modulated and a cyclic prefix is added to the beginning of each OFDM symbol.

At the receiver, the UE receives the signal, eliminates the cyclic prefix, and applies Fast Fourier Transform (FFT) to reconstruct all OFDM symbols in the frame, including PRSs. Assuming tight synchronization between the receiver and base station, the UE correlates the received PRS signal with the known PRS pattern. The UE identifies the peak of the correlation function, which corresponds to the \toa of the signal from the base station. This process is repeated for all BSs from which signals are received.

\subsection{Problem Definition}
\label{sec:methodology}
In this section, we formulate the MAV localization problem in 5G networks combining IMU data, \ie~angular velocity and linear acceleration, with \toa measurements, which provide the radial distance from multiple BSs. These measurements could then be fused to track the MAV motion by the estimation approaches described in the following sections. 

It is worth noting that while recent breakthroughs in machine learning, like Variational Autoencoders (VAEs) and other learning-based approaches, hold promise for localization tasks, this study concentrates on model-based estimation techniques, specifically the ESKF and PGO, for a few key reasons:

\begin{itemize}
      \item \textbf{Data Availability, Offline Learning}
      \begin{itemize}
            \item  Learning-based methods often require large datasets for effective learning, which can be challenging to obtain, especially for unique environments or specific 5G deployments. Although transfer learning could mitigate this issue, it falls outside the scope of our current study due to the additional data needed for fine-tuning and the potential computational burden on resource-constrained drones. As such, we prioritized alternative approaches better suited to our real-time and resource-constrained scenario.
      \end{itemize}

  \item \textbf{Interpretability and Safety:}
  \begin{itemize}
    \item In safety-critical applications like drone navigation, understanding the decision-making process behind localization is paramount. Model-based methods such as ESKF and PGO provide greater transparency in their calculations compared to learning-based methods, which can be black boxes and difficult to interpret. This transparency aligns better with the safety requirements of our application.
  \end{itemize}
\end{itemize}

\subsubsection{Sensing and State Representation for MAV Localization}
We aim to determine the 3D location and orientation of the MAV's body center, which we align with the IMU frame $\Imu$, in the world fixed frame $\World$.  The set of state variables $\States$ contains the pose and velocity of the MAV.  Moreover, we need to estimate the time-variant biases of the IMU's gyroscope $\mathbf{b}^g \in \mathbb{R}^3$ and accelerometer $\bias^a \in \mathbb{R}^3$ to account for the IMU noise drift.

The state vector $\States$ can then be represented as:

\begin{equation}\label{states}
    \States = 
    \begin{bmatrix}
    \mathbf{q}^T & (\mathbf{b}^g)^T & \vel^T & (\bias^a)^T & \tran^T
    \end{bmatrix}^T.
\end{equation}

Here, each transformation $\T = (\R,\tran) \in \SEthree$ is composed of a rotation $\R \in \SOthree$ and a translation $\tran \in \mathbb{R}^3$. This transformation matrix $\T$ effectively transforms the body frame $\Imu$ to the world frame $\World$, in which the base stations (BSs) are positioned. Additionally, the quaternion form of rotation, denoted as $\mathbf{q}$, is related to the rotation matrix $\R$ in the transformation $\T$. 

\subsubsection{IMU Measurements}
IMU measurements are provided w.r.t. the $\Imu$ frame. Our approach involves a 6-axis IMU that measures the linear acceleration ${}_\imu\bar\acc$ and angular velocity  ${}_\imu\bar\rotvel$. The IMU real motion state $\{ {}_\imu\acc, {}_\imu\rotvel\}$ is altered by additive Gaussian white noise $\{\noise^{a}, \noise^{g}\}$ and slowly time-varying biases $\{\bias^a, \bias^g\}$ affecting respectively the accelerometer and gyroscope as defined by the following IMU measurement model \cite{chatfield1997fundamentals, 4209642}:

\begin{equation}\label{eq:imu-model}
\begin{aligned}
  {}_\imu\bar\rotvel
    =&\ {}_\imu\rotvel +\R^T \rotvel^G+ \bias^g + \noise^g, \\
  {}_\imu\bar\acc
   = &\ \R^\trnsp\left( {}_\world\acc-\gravity +2\rotvel^G_\times\vel +[\rotvel^G_\times]^2 \tran\right)\\
    &+ \bias^a + \noise^a \, 
\end{aligned}
\end{equation}

where ${}_\imu\bar\rotvel$ and ${}_\imu\bar\acc$ denote the gyro and accelerometer measurements in the body frame,  and ${}_\world\acc$ is the acceleration of the body frame expressed in the world frame. The terms $\noise^a$ and $\noise^g$ represent the stochastic noise affecting accelerometer and gyroscopic measurement readings, respectively. 

The term $\rotvel^G$ incorporates the effects of the earth's rotation, and the accelerometer measurements include the gravitational acceleration $\gravity$. The operator $\rotvel_\times$ denotes a skew-symmetric matrix, and for $\rotvel = (\rotvel_x, \rotvel_y, \rotvel_z)$, it is defined by:

\begin{equation*}
   \rotvel_\times =  \begin{bmatrix}
0 & -\rotvel_z & \rotvel_y \\
\rotvel_z & 0 & -\rotvel_x \\
-\rotvel_y & \rotvel_x & 0 \\
\end{bmatrix}.
\end{equation*}

\subsubsection{Time of Arrival (\toa) Measurements }
In this context, we leverage \toa measurements denoted as $\bar\rho = (\bar\rho_{1}, \ldots, \bar\rho_{K})$, where $K$ is the total number of 5G BSs. By multiplying the \toa values with the speed of light $c$, we can deduce the distances to each BS, respectively. Let us represent these distances using the vector:
\[
\bar{\mathbf{d}} = (\bar{\mathbf{d}}_{1}, \ldots, \bar{\mathbf{d}}_{K})^T.
\]
This distance is determined by both the dynamic position of the drone and the fixed positions of the 5G BSs (Landmarks), defined by the equation:
\begin{equation}\label{eq:dist_eq}
  \bar{\mathbf{d}}_k  = 
  \mathbf{d}_{k} + \noise^{\text{Dist}}_k =  \| \tran - \mathbf{L}_k\|_2  + \noise^{\text{Dist}}_k, \quad k \in \{1, ..., K\}.
\end{equation}

Here $\bar{\mathbf{d}}_{k}$ denotes the distance measurement between the drone and the $k$-th BS and  $\mathbf{L}_k \in \mathbb{R}^3$ denotes the 3D position of the $k$-th 5G BS. 

Please note that unless explicitly stated, the explicit time index is omitted for brevity and all variables will be interpreted as values at time $t$.

\subsubsection{Dynamic MAV model}

Assuming the MAV has a rigid frame, we apply the dynamics of a rigid body to characterize its motion. The MAV's attached IMU allows us to use IMU measurements as inputs to the dynamic model. This results in the following dynamic model governing IMU states:

\begin{equation}\label{eq:state-model}
    \begin{aligned}
    \dot{\mathbf{q}} &= \frac{1}{2} \Omega\left(\rotvel(t)\right) \mathbf{q}(t),\\
    \dot \bias^g &= \noise^{wg}(t),\\ 
   \dot \vel &= {}_\world\acc, \\
    \dot \bias ^a &= \noise^{wa}(t),\\
    \dot \tran &= \vel,
\end{aligned}
\end{equation}

where $\noise^{wa}$ and $\noise^{wg}$ are random walk noise of IMU accelerometer and gyroscope biases, and  

\begin{align}
\Omega(\rotvel) = \begin{bmatrix}
    -\rotvel_\times & \rotvel \\
    -\rotvel^T  & 0
\end{bmatrix}.
\end{align}

Taking into account the IMU measurement model described in  Eq.~\eqref{eq:imu-model} and applying the expectation operator $\hat{ }$, we obtain the following set of dynamic equations:

\begin{equation}\label{eq:post_error_state}
\begin{aligned}
    \dot{\hat{\mathbf{q}}} &= \frac{1}{2} \Omega\left(\hat\rotvel(t)\right) \hat{\mathbf{q}}(t),\\
    \dot {\hat\bias}^g &= \bf{0},\\
    \dot{\hat\vel} &= \hat\R^T\hat\acc + \gravity - 2[{\rotvel^G}_\times]\hat\vel - [\rotvel^G_\times]^2 \hat\tran,   \\
    \dot {\hat\bias} ^a &= \bf{0},\\
    \dot {\hat\tran} &= \hat\vel,
\end{aligned}
\end{equation}

with $\hat \acc  = {}_\imu\bar\acc - \hat \bias ^a$ and $\hat \rotvel = {}_\imu\bar\rotvel - \hat \bias_g - \hat\R^T \rotvel^G$.
Note that we consider $\World$ as an inertial frame of reference, neglecting the effect of the Earth's rotation, i.e., $\rotvel^G = \bf{0}$.

\subsection{Error State Kalman Filter for MAV Localization (Indirect Method)}\label{sec:eskf}

The Extended Kalman filter (EKF) is a widely employed technique for estimating and tracking system states, particularly for non-linear system dynamics or measurement models. The EKF operates by linearizing the system dynamics and the observation (measurement) model around the current state estimate. Although the EKF performs accurately enough in many situations, its efficacy may diminish in highly nonlinear systems.

One effective approach is to utilize the error-state Kalman filter (ESKF) method for enhanced precision and confidence in estimations, as elucidated by \cite{panich2010indirect}. ESKF entails approximating the error state, representing the disparity between the true and estimated states, rather than directly estimating the states themselves.  The rationale behind adopting an ESKF, also known as the indirect method, is that errors are typically smaller and exhibit more linearity than the states, making them well-suited for linear Gaussian filtering. This enables the ESKF to refine estimations with greater precision.  This process unfolds in two main steps, namely prediction and update.

During the prediction stage, the system state is forecasted using integration, disregarding minor disturbances and noise. However, this straightforward projection can result in the gradual accumulation of errors due to noise and disturbances. The ESKF simultaneously calculates a Gaussian estimate of the error state's distribution while integrating the nominal state to address these inaccuracies. This dual process refines our real-time understanding of the evolving errors. By identifying small, subtle signals within the system's behavior, the ESKF enhances state estimation accuracy. 

The update step is triggered by the reception of the new measurements, a less frequent occurrence than prediction. These measurements serve to expose accumulated errors. Exploiting this new information, the ESKF refines the error state estimate, enhancing the comprehension of its distribution and contributing to higher accuracy. The error state is then augmented with the estimated state, enabling necessary adjustments. While the error state is reset to zero, its uncertainty is updated.

Table \ref{tab:ekf-vs-eskf} presents the key differences between the EKF and ESKF, highlighting the trade-off between accuracy and computational complexity. In summary, while  EKF directly estimates states and linearizes around the current estimate, leading to lower computational complexity, its accuracy may be compromised in highly non-linear systems. The ESKF, on the other hand, tackles this challenge by estimating the error between true and estimated states and linearizing around this error state. This approach achieves higher accuracy in non-linear scenarios, although at a slightly increased computational cost due to additional calculations and memory requirements. In both filters, the prediction stage utilizes linearized dynamics to forecast the state (ESKF predicts both error and nominal state), while the update stage refines the estimate using new measurements (ESKF updates error before refining the full state).

\begin{table*}
  \centering

  \caption{Comparison of Extended Kalman Filter (EKF) and Error State Kalman Filter (ESKF)}
  \label{tab:ekf-vs-eskf}
  \setlength{\extrarowheight}{0.5em} 
  \resizebox{1\linewidth}{!}{
    \begin{tabular}  { M{0.18\textwidth} M{0.4 \textwidth} M{0.4 \textwidth}}

\Xhline{3\arrayrulewidth}
\bf Feature & \bf EKF & \bf ESKF   \\
\Xhline{1\arrayrulewidth}
\rowcolor{mygray} State Representation & Actual system states & Error between true and estimated states \\
                  Linearization Point & Current state estimate  & Error state \\
                  
\rowcolor{mygray} Suitability for Non-linearity & Less accurate & More accurate \\
                  Computational Complexity & Lower & Slightly higher \\

\rowcolor{mygray} Prediction Stage & Predicts system state based on linearized dynamics & Predicts error state evolution \& nominal state
 \\
                  Update Stage & Updates state estimate directly	 & Updates error state estimate \& refines state through error augmentation \\

\Xhline{3\arrayrulewidth}

\end{tabular}
}
\end{table*}

\subsubsection{Prediction Step}
 Having the states $\States$ defined in  Eq.~\eqref{states}, the error state is defined as per \cite{4209642}:
 
\begin{gather}\label{eq:err-state}
    \delta\States = 
    \begin{bmatrix}
    \delta \theta^T & (\delta\bias^g)^T & \delta\vel ^T & (\delta\bias^a )^T & \delta\tran^T
    \end{bmatrix}^T.
\end{gather}

Herein, $\delta x = x- \hat x$, where $x$ represents the true value, and $\hat x$ is the estimated value. However, the quaternion error is defined as $ \mathbf{q} = \delta \mathbf{q} \otimes \hat {\mathbf{q}}$, with $\mathbf{q}$ being a real value and $\hat{\mathbf{q}}$ as the estimated one. To effectively represent slight rotational errors, we adopt the following approximation:

\begin{gather}
    \delta \mathbf{q} \simeq \begin{pmatrix}
        \frac{1}{2}\delta \theta^T & 1
    \end{pmatrix}^T.
\end{gather}

This approximation serves to simplify the representation of minor rotational errors within quaternion-based state estimation.

Note that we explicitly estimate the IMU biases states within the overall state vector($\delta\States$). This allows the filter to learn and compensate for these constant offsets over time. In the error state equation, bias state dynamics would be modeled as:
$\dot{\delta\bias^g} \approx 0$ and $\dot{\delta\bias^a} \approx 0$ (assuming random walk noise is negligible).

Employing the minimal representation provided by $\delta \theta$, it reduces computational complexity while simultaneously maintaining accuracy.

Based on  Eq.~\eqref{eq:post_error_state} and  Eq.~\eqref{eq:err-state}, the error state equation can be expressed in the following form:

\begin{gather}\label{eq:dynamic_error_state}
    \delta\dot {\States} = \F \delta \States +  \G \noise^{\text{IMU}},
\end{gather}
where $\noise^{\text{IMU}} = \begin{bmatrix}
    (\noise^g)^T & (\noise^{wg})^T & (\noise^a)^T & (\noise^{wa})^T
\end{bmatrix}^T$, $\F$, and $\G$ are given as follows:

\begin{gather*}
        \F = \begin{bmatrix}
        -[\hat\rotvel_\times] & -I_3  & 0_{3\times3} & 0_{3\times3} & 0_{3\times3} \\
        0_{3\times3} & 0_{3\times3} & 0_{3\times3} & 0_{3\times3} & 0_{3\times3} \\
        -\hat\R^T[\hat\acc _\times] & 0_{3\times3} & -2[\rotvel^G_\times] & -\hat\R^T & -[\rotvel^G _\times]^2 \\
        0_{3\times3} & 0_{3\times3} & 0_{3\times3} & 0_{3\times3} & 0_{3\times3} \\
        0_{3\times3} & 0_{3\times3} & I_3 & 0_{3\times3} & 0_{3\times3} \\
    \end{bmatrix},
\end{gather*}

and \begin{gather*}
        \G = \begin{bmatrix}
        -I_3 & 0_{3\times3} & 0_{3\times3} & 0_{3\times3} \\
        0_{3\times3} & I_3 & 0_{3\times3} & 0_{3\times3} \\
        0_{3\times3} & 0_{3\times3} & -\hat\R^T & 0_{3\times3} \\
        0_{3\times3} & 0_{3\times3} & 0_{3\times3} & I_3 \\
        0_{3\times3} & 0_{3\times3} & 0_{3\times3} & 0_{3\times3} \\
    \end{bmatrix},
\end{gather*}
where $I_3 \in \mathbb{R}^{3 \times 3}$ denotes the Identity matrix.

In the discrete implementation, the IMU captures measurements at discrete time intervals, namely $\bar\rotvel_t$ and $\bar\acc_t$, which are then employed to propagate the state and covariance matrix. The ESKF is initialized with the initial state estimate and covariance matrix. Upon the arrival of a fresh IMU measurement, the IMU-derived state estimate undergoes propagation through a  numerical integration, effectively accounting for the dynamic model of the system as specified by  Eq.~\eqref{eq:post_error_state}. The covariance matrix is also propagated using a numerical integration of the Lyapunov equation: 
\begin{gather}\label{eq:lyapunov}
   \dot \Pcov  = \F \Pcov + \Pcov \F^T +\G \mathbf{Q}^{\text{IMU}}\G^T.
\end{gather}

Note that the IMU noise characteristics are represented by a noise covariance matrix ($\mathbf{Q}^{IMU}$) tailored to the specific sensor used. This matrix captures the variances and correlations of different noise sources (gyroscope noise, accelerometer noise, etc.).
Accurate estimation of $\mathbf{Q}^{IMU}$, whether through empirical measurements or manufacturer specifications, is crucial to precise noise modeling. 
The $\mathbf{G}$ matrix in the error state equation \eqref{eq:dynamic_error_state} acts as a gain matrix, mapping IMU noise directly into the error state dynamics. Each non-zero entry in $\mathbf{G}$ indicates how a specific noise component influences a particular state error. For example, gyroscope noise primarily affects the attitude error, while accelerometer noise impacts both attitude and velocity errors. Furthermore, in the propagation of the covariance matrix, the process noise term ($\G \mathbf{Q}^{IMU}\G^T$) in the Lyapunov equation \eqref{eq:lyapunov} drives the growth of the state covariance matrix ($\Pcov$).
As the covariance matrix increases, it reflects the increasing uncertainty in the state estimates due to IMU noise accumulation

\subsubsection{Update Step}
Upon receiving Time of Arrival (ToA) measurements, we proceed with the update stage. Given  Eq.~\eqref{eq:dist_eq}, we define the measurement function $h(\States)$ as follows:

\begin{gather*}
    h(\States) = \begin{bmatrix}
    \| \tran - \mathbf{L}_1\|_2 \\
    \vdots\\
    \| \tran - \mathbf{L}_K\|_2 \
\end{bmatrix} \in \mathbb{R}^{K \times 1}.
\end{gather*} 

Now the residual $\residual^\rho \in \mathbb{R}^{K\times 1}$ is formulated as:
\begin{equation}
\begin{aligned}
   \residual^\rho = \: & \bar{\mathbf{d}} - h(\hat\States) =  h(\States) - h(\hat\States)+ \noise^{\text{Dist}}  \\ =\: & h(\hat\States+\delta\States) -h(\hat\States)+ \noise^{\text{Dist}}\\
   \approx \: & \mathbf{H} \delta\States+ \noise^{\text{Dist}},
\end{aligned}
\end{equation}
where $\noise^{\text{Dist}}$ represents the vector of noise associated with the distance measurements and the measurement Jacobian matrix $\mathbf{H}$ is defined by:
\begin{equation}
    \begin{aligned}
        \mathbf{H} = & \frac{\partial \residual^\rho}{\partial(\delta\States)}|_{ \hat\States} \\
        = & \begin{bmatrix}
       0_{1\times 12} & \frac{\hat\tran_x -L_{1x}}{d_1} & \frac{\hat\tran_y -L_{1y}}{d_1} & \frac{\hat\tran_z -L_{1z}}{d_1}\\
       \vdots & \vdots & \vdots & \vdots  \\
       0_{1\times 12} & \frac{\hat\tran_x -L_{Kx}}{d_K} & \frac{\hat\tran_y -L_{Ky}}{d_K} & \frac{\hat\tran_z -L_{Kz}}{d_K}
   \end{bmatrix}_{k \times 15}
\end{aligned}
\end{equation}

Here  $L_k = (L_{kx}, L_{ky}, L_{kz})$, for $k \in {1, ..., K}$ represent the position of the k$th$ base station. 

The update rules at time $t$ will be:
\begin{align}
    \mathbf{K}_t = & \mathbf{P}_{t|t-1} \mathbf{H}^T  \left(\mathbf{H} \mathbf{P}_{t|t-1} \mathbf{H}^T + \mathbf{R} \right)^{-1},\\
    \delta\hat\States_{t|t} =&  \Kgain_t \residual^\rho_t,\\
    \Pcov_{t|t} = & (\mathbf{I} - \Kgain_t \mathbf{H}) \Pcov_{t|t-1}.
\end{align}

In these equations, $\mathbf{R}$ represents the covariance matrix of the distance measurement noise. The process involves calculating the Kalman Gain $\mathbf{K}_t$, updating the error state estimate $\delta\hat\States_{t|t}$ and the error covariance matrix. Finally, $\delta\hat\States_{t|t}$ is appropriately incorporated into the predicted state $\hat\States_{t|t-1}$ based on the specific calculations for each component of the state vector. This step refines the estimated state using ToA measurements and their associated uncertainties.

The ESKF algorithm's core steps are outlined in the pseudocode presented in \autoref{alg:eskf}, serving as an overview of the ESKF implementation, helping to grasp the algorithm's logical flow and key components.

\begin{algorithm}[!h]
    \SetAlgoNlRelativeSize{-1}
    \caption{ESKF-based Localization}
    \label{alg:eskf}

    \KwIn{IMU and ToA measurements}
    \KwResult{state estimate $\hat{\States}$}

    Initialize: $\hat{\States}_0$, $\Pcov$ \;
    \While{new IMU measurement received}
    {
          \BlankLine
         \textbf{\underline{Prediction Stage:}} \\
        Propagate state using Runge-Kutta integration\;
        Propagate covariance matrix using Runge-Kutta integration\;
        
        \If{new ToA measurements received}
        {
            \BlankLine
            \textbf{\underline{Update Stage:}} \\
            
            Obtain distance measurements $\mathbf{\bar{d}}_t$\;
            Calculate residual $\residual^\rho_t \leftarrow \mathbf{\bar{d}}_t - h(\hat{\States}_{t|t-1})$\;
            Calculate Jacobian matrix $\mathbf{H}_t$ for measurement model\;
            Calculate Kalman Gain $\mathbf{K}_t$\;
            Update error state estimate $\delta\hat{\States}_{t|t} \leftarrow \mathbf{K}_t \residual^\rho_t$\;
            Update covariance matrix $\Pcov_{t|t} \leftarrow (\mathbf{I} - \mathbf{K}_t \mathbf{H}_t) \Pcov_{t|t-1}$\;
            Update state estimate $\hat{\States}_{t|t} \leftarrow \hat{\States}_{t|t-1} \oplus \delta\hat{\States}_{t|t}$\;
        }
    }
\end{algorithm}


\subsection{Pose Graph Optimization (PGO)}\label{sec:pgo}
\begin{figure}[!ht]
    \centering
    \includegraphics[width = 0.98\linewidth]{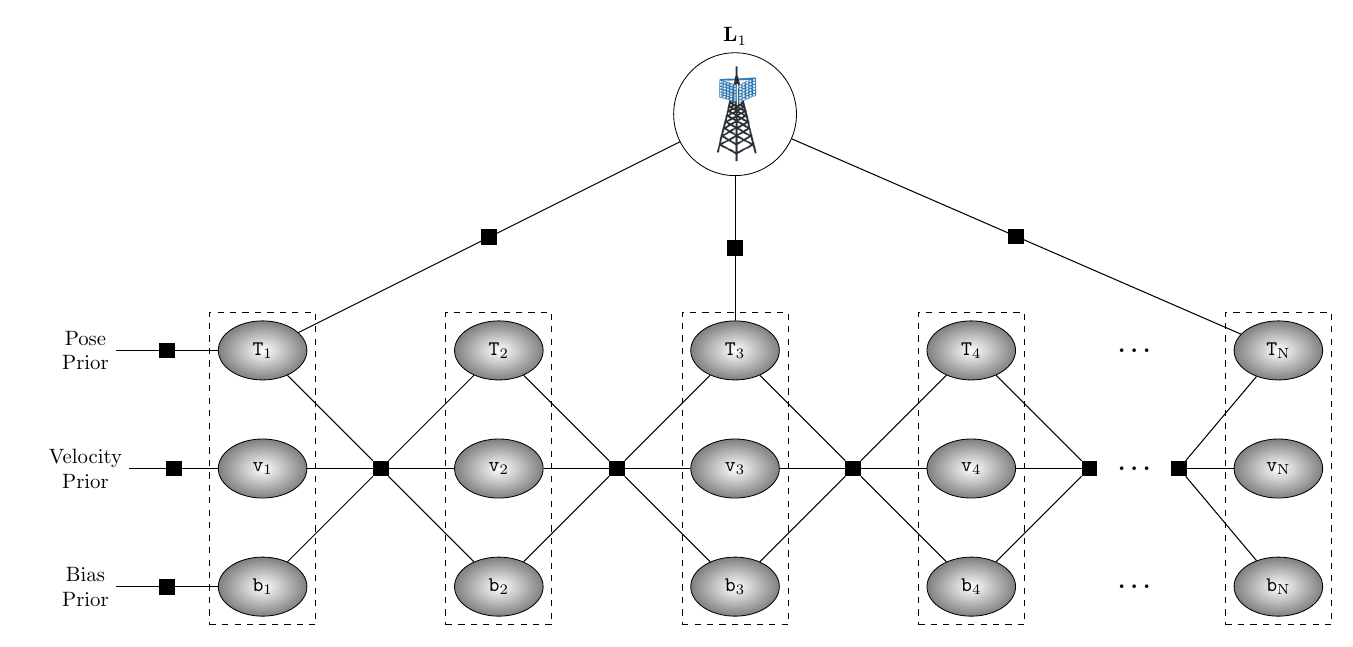}
 \caption{The figure visualizes the structure of the factor graph used to optimize the variables, represented by circles, by relating them through factors, represented by squares. The nodes $\T_t$ incorporate the 6DoF pose variables, $\mathtt{v}_t$ nodes encapsulate the velocity variables, and $\mathtt{b}_t$ nodes denote the bias variables, encompassing biases from both gyroscopes and accelerometers. IMU pre-integration factors connect all of these nodes. \toa measurements create range factors between robot pose nodes and BSs position nodes, with a single instance, $\mathbf{L}_1$, visualized here to enhance graph clarity.  Prior factors, namely prior pose, prior velocity, and prior bias, are connected to the respective nodes $\T_1$, $\mathtt{v}_1$, and $\mathtt{b}_{1}$ to constrain them with their initial values in the trajectory.}
    \label{fig:graph}
\end{figure}

As a second alternative method to estimate the drone's 6DoF pose we also leverage graph-based optimization techniques. This method models the relationships between the pose variables based on sensor measurements and then performs the estimation using least squares optimization. For that purpose, a factor graph model~\cite{factor_graphs_for_robot_perception} is created to abstract the problem. Factor graphs are bipartite graphs with two types of nodes, namely variables and factors, where factors represent functions on subsets of the variables. Edges in the factor graph between a factor and a set of variables indicate that the factor depends only on those variables. In our case, the variables are the state variables that should be estimated at certain instants and the factors correspond to the likelihood of the adjacent state variable nodes given the related measurements, see the structure of the resulting factor graph in Fig.~\ref{fig:graph}. The figure illustrates a factor graph used for optimizing variables, with circles representing variables and squares representing factors. Nodes $T_t$ represent pose variables, $v_t$ represent velocity, and $b_t$ represent bias. IMU pre-integration factors connect these nodes, while ToA measurements create range factors. Prior factors constrain initial values. 

Therefore, the factor graph can be used to specify the joint posterior probability density $p(\mathcal{X}|\mathcal{Z})$ of the whole set of state variables $\mathcal{X}$ and the whole set of measurements $\mathcal{Z}$ as the product of all factors $f_i(\mathcal{X}_i)$ in the graph

\begin{equation}
\label{eq:fg}
    p(\mathcal{X}|\mathcal{Z})  \propto \prod_{\forall f_i\in \mathcal{F}} f_i(\mathcal{X}_i) \, ,
\end{equation}
where $\mathcal{X}_i$ is the set of state variable nodes adjacent to the factor $f_i(\mathcal{X}_i)$ and $\mathcal{F}$ is the set of all factors in the graph. Herein, the measurements are no longer explicitly represented but simply become a fixed parameter in the corresponding factor. This factorization will be exploited for pose estimation as described in the following.  

\subsubsection{IMU Factor}
Due to the IMU's higher sampling frequency than other sensors, it typically captures multiple measurements between two \toa instances. The IMU factor is constructed utilizing a  \textit{preintegrated measurement}~\cite{imu_preintegration} constraining the relative motion increments. Especially, we obtain the condensed measurements $\preintRmeas_{ij}$ of rotation, $\preintPmeas_{ij}$ of position, and $\preintVmeas_{ij}$ of velocity by integrating multiple IMU readings $\{{}_\imu\bar\acc_t, {}_\imu\bar\rotvel_t : \forall t \in [t_i, t_j]\}$. So, we can define the residual terms $\residual$ for the states $\{\R_{ij}, \tran_{ij}, \vel_{ij}\}$:
\begin{align}
  \residual_{ij}^{\R}
    = &\ \textstyle \logmap\left( \preintRmeas_{ij}^\trnsp\
     \R_i^\trnsp \R_j  \right)\, ,\\                  
  \residual_{ij}^{\tran}
     = &\ \textstyle  \R_i^\trnsp \big( \tran_j-\tran_i - \vel_i\Delta t_{ij}
    - \frac{1}{2}\gravity \Delta t_{ij}^2\big)  -\preintPmeas_{ij}\, ,\\    
  \residual_{ij}^{\vel} 
     = &\ \textstyle \R_i^\trnsp \left(\vel_j-\vel_i - \gravity\Delta t_{ij}\right)-\preintVmeas_{ij} \, ,
\end{align} 
where $\Delta t_{ij} = t_j - t_i$ is the total time interval with $t_i < t_j$. Also, $\logmap : \SOthree \rightarrow \Real^3$ defines the logarithm map that associates elements of the rotation manifold $\SOthree$ to vectors on the Euclidean tangent space $\Real^3$ representing rotation increments.
Regarding the biases, the total residual $\residual_{ij}^{\bias}$ between time $t_i$ and $t_j$ is set as follows: 
\begin{equation}
    \residual_{ij}^{\bias} =\ \bias^{g}_j - \bias^{g}_i + \bias^{a}_j - \bias^{a}_i  \, .
\end{equation}

\subsubsection{\toa Range Factor}

By multiplying the estimated \toa values $\rho_{sk}$ by the speed of light $c$, \ie~$d_{sk} = \rho_{sk} \cdot c$, 
we obtain $\mathrm{K}$ metric distance measurements $d_{sk}\in \Real$ of the drone to the $k$-th BS at location $\mathbf{L}_k$ at time $t_s$. Notably, we explicitly express the possibility of having fewer \toa measurements than the number of tracked poses. The residual $\residual_{sk}^{\rho}$ of the \toa factor at time $t_s$ with the BS $\mathbf{L}_k$ is defined as: 
\begin{equation}
  \residual_{sk}^{\rho} = d_{sk} - \| \tran_s  - \mathbf{L}_k\|_2 \, .
\end{equation}

\subsubsection{Optimization}

The pose graph optimization problem is formulated as Maximum a Posteriori (MAP) estimation that involves finding the state $\mathcal{X}^*$ that maximizes the posterior:

\begin{equation}
\label{eq:map0}
    \mathcal{X}^* = \argmaxA_{\mathcal{X}}  p(\mathcal{X}|\mathcal{Z}) .
\end{equation}
Considering the proportional relationship in  Eq.~\eqref{eq:fg},  Eq.~\eqref{eq:map0} is equivalent to the maximization of the product of all factors in the factor graph:
\begin{equation}
\label{eq:map}
    \mathcal{X}^* = \argmaxA_{\mathcal{X}} \prod_{\forall f_i\in \mathcal{F}} f_i(\mathcal{X}_i)\, .
\end{equation}
In our application, the factor graph evolves with time, where we consider $t=\{1,\ldots,N\}$ instants and $\mathcal{X}_t$ is the state at instant $t$. Therefore, the overall set of states is $\mathcal{X} = \{\mathcal{X}_1,\ldots, \mathcal{X}_N\}$ starting with a given initial state $\mathcal{X}_0$. The factors are likelihoods derived from the respective previously described measurements, assumed to be corrupted by zero-mean, normally distributed noise. Now we denote with $f_{ti}$ the factors between state $\mathcal{X}_t$ and the state $\mathcal{X}_{t+1}$ and, if \toa measurements are available, the factors between state $\mathcal{X}_t$ and the respective BSs. Taking the negative log of  Eq.~\eqref{eq:map} finally leads to the minimization of the sum of the respective residuals in the following form: 

\begin{equation}
\mathcal{X}^* =\ \argminA_{\mathcal{X}} \left\|\residual_0\right\|^2_{ \mathbf{\covprior}_0} + \sum^\mathrm{N-1}_{t=1} \sum^{}_{\forall f_{ti}\in{\mathcal{F}}}
\left\|\residual_{f_{ti}}\right\|^2_{ \mathbf{\covprior}_{f_{ti}}} \, ,
\end{equation}
where $\left \| \mathbf{r} \right\|_\mathbf{\covprior}^2 = \mathbf{r}^\trnsp \mathbf{\covprior}^{-1}{\mathbf{r}}$ is the squared Mahalanobis norm, and $\residual_{f_{ti}}$ are the residual functions related to the aforementioned factors $f_{ti}$ with covariance matrix $\mathbf{\covprior}_{f_{ti}}$. We denote with $\residual_0$ the residual derived from the prior on the initial pose with $\mathbf{\covprior}_0$ being its covariance matrix.

To efficiently solve the MAP optimization problem, we utilize the iSAM2 (Incremental Smoothing and Mapping 2) iterative optimization algorithm~\cite{ISAM2} implemented in GTSAM~\cite{gtsam}. This algorithm can automatically identify the variables that require linearization at each step, and it enables us to keep our graph solution updated while adding new nodes without experiencing memory overload.

iSAM2 effectively leverages a Bayesian tree structure to incorporate historical data during optimization, but its performance can be impacted by the growing tree depth for extended temporal horizons. In scenarios where both long-term memory retention and real-time performance are critical, \cite{tazaki2022spanning} offers a promising solution that is based on a spanning tree-based hierarchy.  This method leverages a simplified spanning tree representation of the pose graph, reducing complexity. It also employs a coarse-to-fine optimization strategy, achieving faster convergence by first optimizing on supernodes formed by connected components and then refining within each supernode using the original measurements. While this approach introduces a slight accuracy approximation, its reduced complexity and potential for real-time performance make it a valuable option for specific applications.

Within \autoref{alg:gtsam}, we have encapsulated the essence of graph-based state estimation in a streamlined manner.

\begin{algorithm}
\SetAlgoLined
\KwIn{IMU and ToA measurements}
\KwResult{Optimized state estimates $\mathcal{X}^*$}
Initialize: $\mathcal{X}_0^* = \mathcal{X}_0$\;
\While{true}{
    \For{each fixed time interval}{
        Accumulate IMU measurements\;
        preintegrate accumulated IMU measurements\;
        Add a new node\;
        Create an IMU factor linking the new node to the previous one\;
        \If{New ToA measurement arrives}{
            Determine the temporally nearest node to the ToA measurement.\;
            Incorporate a range factor between the node and the relevant landmark\;
        }
        Perform optimization using iSAM2\;
        Update the state estimates $\mathcal{X}^*$\;
    }
}
\caption{Graph-Based Localization}
\label{alg:gtsam}
\end{algorithm}

\section{Evaluation and Results}
\label{sec:evaluation}
In order to evaluate the two derived approaches for pose estimation, we intend to apply them as far as possible to real experimental data. For this purpose we selected the EuRoC MAV dataset~\cite{Burri2016} that comprises vision data and inertial sensor measurements as well as the related ground truth pose data collected during a number of MAV indoor flight sequences. The dataset is widely used in the literature as a benchmark for the evaluation of visual-inertial odometry or SLAM solutions. However, while it provides the inertial measurements for our evaluation, it does so far not include the necessary 5G ToA data. Therefore, we had to augment the EuRoc MAV dataset with such suitable data that could either be generated experimentally or via simulations. However, an experimental generation of these 5G \toa data would not only require an exact repetition of the flights with the same equipment and configurations, but also the setup of a configurable 5G indoor network and an on-board 5G receiver with ToA measurements. While such 5G networks (e.g., 5G femtocells) are foreseen for the near future by many mobile communications equipment suppliers, it is currently still a major challenge to acquire them commercially and install them in an academic lab environment.

Therefore, we decided to generate the missing 5G ToA data via a suitable simulation environment. In the following, we describe in more detail our approach to generate such simulated yet highly realistic data and finally present our assessment based on the augmented EuRoC MAV dataset.

\subsection{Augmenting the EuRoC MAV dataset with simulated 5G \toa data}
\begin{figure*}[htpb]
    \centering
    \includegraphics[width = 0.8\textwidth]{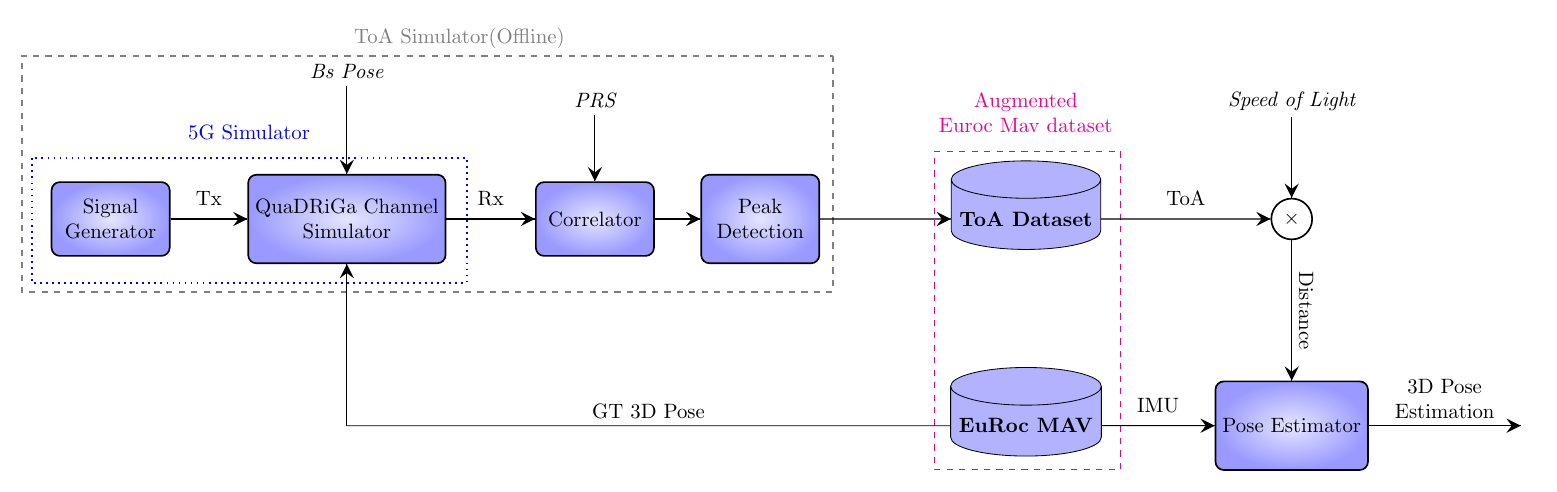}
    \caption{System Architecture for Augmentation of EuRoC MAV Dataset with 5G Simulation.}
    \label{fig:arch}
\end{figure*}

Fig.~\ref{fig:arch}  shows the overall structure of the simulation environment with its core components and their interaction for the augmentation of the EuRoC MAV dataset. In addition, it shows the connection of the resulting dataset with the pose estimators for the assessment.
Herein, the main components are:
\begin{itemize}
    \item {\bf EuRoC MAV dataset:} provides inertial measurements and ground truth pose data from real MAV flights.
    \item {\bf Signal generation:} Generates the 5G signals as emitted from a number of simulated BSs placed at defined locations. 
    \item {\bf QuaDRiGa channel simulation:} Simulates the propagation of the 5G signals from the BSs with their respective fixed poses through the environment to a supposed flying MAV with its current pose provided by the EuRoC MAV ground truth data. 
    \item {\bf \toa Calculation:} Calculates the current \toa measurements as provided by a supposed MAV on-board 5G receiver by analyzing the received 5G signal, and stores them as a dataset.
    \item {\bf Pose estimator:} Estimates the pose of the MAV using the stored simulated \toa measurements and the inertial measurement from the EuRoC MAV dataset using one of the two pose estimators (i.e., ESKF or PGO) as derived in Sections \ref{sec:pgo} and \ref{sec:eskf}, respectively.
\end{itemize}

The EuRoC MAV dataset~\cite{Burri2016} was collected by an indoor flying MAV equipped with a stereo-camera module capturing images at 20 Hz and a calibrated IMU providing inertial measurements at a rate of 200 Hz. The dataset also containes the MAV's position and orientation data obtained through the Vicon motion capture system as ground truth, recording the full 6DoF at approximately 100 Hz. The full set of calibrated rigid transformations between sensors and the Vicon system is also given. The EuRoC MAV dataset comprises multiple flight sequences while our study examines all six sequences from the Vicon Room 1 and Vicon Room 2 datasets.

To estimate the distance to 5G BSs, we generate the 5G signal, including PRS and Physical Downlink Shared Channel (PDSCH) resources transmitted by each BS. We use the MATLAB 5G Toolbox to generate resource blocks for these signals, setting the transmit power to 0 dBm (1 mW) and the SNR to 10 dB, which is a conservative value, as real-world scenarios often have higher SNR levels.

We then employ the QuaDRiGa (quasi-deterministic radio channel generator) channel simulator~\cite{jaeckel2017quadriga} to create an impulse response that emulates the wireless channel characteristics based on specific network configurations, given receiver and transmitter positions, orientations, velocities, and the chosen indoor environment. QuaDRiGa is a realistic channel simulator that has been validated through extensive measurements. For the MAV trajectories we have used the available 6DoF ground truth poses of the MAV provided by the EuRoC MAV dataset, from which we compute the required velocity considering the translation vectors between two time-consecutive poses. We virtually place two up to five fictitious BSs in the room where the trajectory is recorded. The positions of the BSs in the EuRoC MAV Vicon system's coordinate frame are $\mathrm{BS}_1 = (-10, -7, 2)$, $\mathrm{BS}_2 = (7, 13, 3)$, $\mathrm{BS}_3 = (25, -35, 4)$, $\mathrm{BS}_4 = (-6, 9, 5)$, $\mathrm{BS}_5 = (-4, -14, 6)$. These values are used to initialize the corresponding state variables of the optimization problem with a small covariance.  

We consider three QuaDRiGa channel simulator scenarios, each operating at different frequencies, to simulate the wireless communication environment realistically: QuaDRiGa-Industrial-LOS for 5 GHz, 3GPP-38.901-Indoor-LOS for 28 GHz, and mmMAGIC-Indoor-LOS for 78 GHz.

\begin{itemize}
    \item {\bf QuaDRiGa\_Industrial\_LOS~\cite{Jaeckel2019}}: This scenario is designed to replicate a line-of-sight (LOS) environment for industrial applications. The simulation is optimized for frequencies ranging from 2 to 6 GHz and aims to capture the radio propagation behavior typically observed in automation industry halls. The scenario is validated through measurements conducted across five factory halls within Siemens' Nuremberg operational premises.

    \item {\bf 3GPP\_38.901\_Indoor\_LOS~\cite{3GPP2018}}: This scenario simulates an indoor environment with a 0.5-100 GHz frequency in LOS conditions. It aims to encompass various indoor deployment scenarios, such as those commonly found in office buildings and shopping centers. These indoor settings typically consist of open cubicle spaces, private enclosed offices, common areas, corridors, etc.

    \item {\bf mmMAGIC\_Indoor\_LOS~\cite{CarneirodeSouza2022}}: This is designed specifically for frequencies in the range of 6-100 GHz and indoor scenarios with LOS like traditional enclosed offices, semi-closed offices (cubicle areas), and open offices.

\end{itemize}

 We also assumed that both the receiver and all transmitters used omnidirectional antennas. Other configurations are detailed in Table~\ref{tab:configs}. It includes information such as frequency band, bandwidth, subcarrier spacing, number of resource blocks (RBs), comb size, signal-to-noise ratio (SNR), and cyclic prefix type for each 5G scenario. 
 \begin{table*}[h!]
    \centering
    \caption{5G system configurations.}
    \label{tab:configs}
    \resizebox{0.8\linewidth}{!}{
    \begin{NiceTabular}{M{4.2cm}M{1.3cm}M{1cm}M{3cm}M{1.5cm}M{1cm}M{1cm}M{1.5cm}}[hvlines]
\CodeBefore
      \Body
        \Xhline{3\arrayrulewidth}
    \bf 5G Sim. Scenario & \bf  Freq (GHz) & \bf Bw (MHz)  & \bf Subcarrier Spacing (KHz) & \bf number of RBs & \bf comb size & \bf SNR  & \bf cyclic prefix\\
    \grayrow QuaDRiGa-Industrial-LOS & 5 (FR1) & 100  & 30 & 275  & 6 & 10 dB & normal \\
    \Xhline{\arrayrulewidth} 
     3GPP-38.901-Indoor-LOS & 28(FR2)& 200  & 60 & 275  & 6 & 10 dB & normal \\
    \Xhline{1\arrayrulewidth}
    \grayrow mmMAGIC-Indoor-LOS & 78(FR2)& 400  & 120  & 275 & 6 & 10 dB & normal \\
    \Xhline{3\arrayrulewidth}

    \end{NiceTabular}
    }
\end{table*}

\begin{figure}
    \centering
    \includegraphics[width=0.7\textwidth]{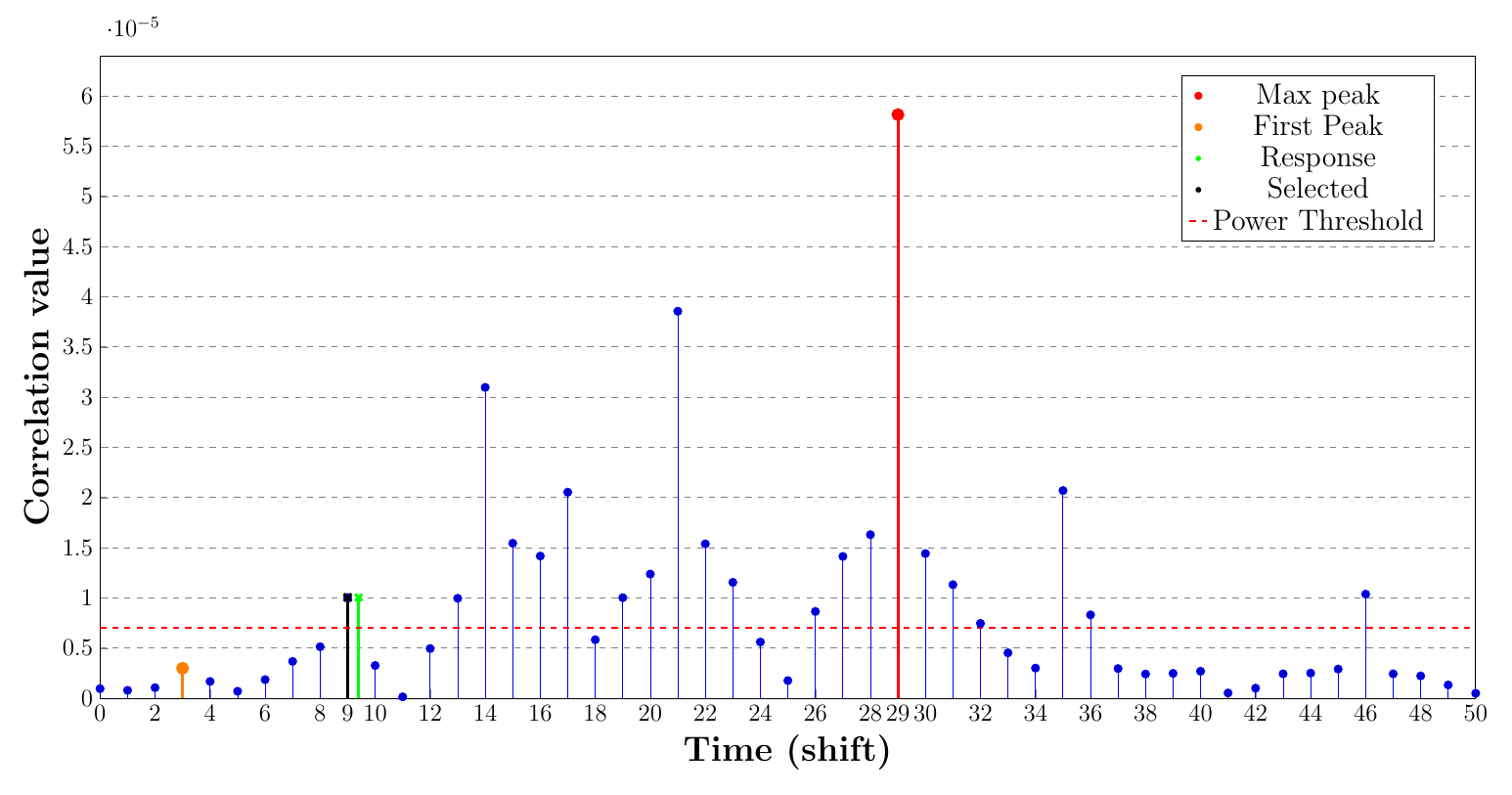}
    \caption{PRS Correlation Profile}
    \label{fig:corr_profile}
\end{figure}

We finally convolve the transmitted signal with the impulse response to replicate the effects of the transmission environment, generating the received signal. We generate the received signal at the receiver every 0.2 seconds, which enables us to calculate the \toa at a frequency of 5 Hz. We then correlate the received signal with the transmitter's PRS pattern and calculate the delay by analyzing the correlation profile. Typically, the initial or highest peak is considered as the response. However, this approach can be compromised by noise or when the LOS coefficient is weaker than the multipath coefficient. To address this, we set a threshold to eliminate values below it and choose the first peak above the threshold as the response.  We have empirically determined that a threshold of 0.2 is optimal for \toa estimation with the generated data. An example of a correlation profile in the simulation is displayed in~Fig.~\ref{fig:corr_profile}. The response shown in green is neither the first nor the maximum peak. Still, a suitable threshold allowed the selection of the first peak as the response. \autoref{tab:stats} gives the statistics of the error in the resulting estimated distance to each BS. 

\begin{table*}[!h]
\setlength\tabcolsep{3.3pt}
\renewcommand{\arraystretch}{1.2}
\centering
  \caption{Error statistics of the estimated \toa distance to BSs for different scenarios and given in meters.}
  \label{tab:stats}
\resizebox{0.9\linewidth}{!}{
\begin{NiceTabular}{M{1.4cm}M{5.5cm}M{1.5cm}M{1.5cm}M{1.5cm}M{1.5cm}M{1.5cm}M{1.5cm}}[hvlines]
\CodeBefore
    \foreach \row in {2, 4, ..., 36} {
      \rectanglecolor{mygray}{\row-3}{\row-8}
      }
\Body
\Xhline{3\arrayrulewidth}
    \bf{Dataset}   & \bf{5G Sim. Scenario}   & \bf Statistic & \bf \toa\#1   & \bf \toa\#2   & \bf \toa\#3   & \bf \toa\#4   & \bf \toa\#5   \\

\Block{6-1}{V101}&\Block{2-1}{QuaDRiGa-Industrial-LOS}&  Mean  &  0.129 &  -0.045 &  0.006 &  -0.081 &  -0.023\\
 &  &  Std. & 0.568 & 0.81 & 0.763 & 0.872 & 0.718\\

 & \Block{2-1}{3GPP-38.901-Indoor-LOS}&  Mean  &  -0.024 &  -0.021 &  -0.059 &  0.041 &  -0.06\\
 &  &  Std. & 0.344 & 0.368 & 0.352 & 0.394 & 0.369\\

 & \Block{2-1}{mmMAGIC-Indoor-LOS}&  Mean  &  0.002 &  0.01 &  -0.008 &  0.003 &  -0.01\\
 &  &  Std. & 0.185 & 0.171 & 0.173 & 0.159 & 0.176\\
\Xhline{4\arrayrulewidth}

\Block{6-1}{V102}&\Block{2-1}{QuaDRiGa-Industrial-LOS}&  Mean  &  0.16 &  -0.033 &  -0.135 &  -0.129 &  -0.156\\
 &  &  Std. & 0.645 & 0.874 & 0.722 & 0.739 & 0.677\\

 & \Block{2-1}{3GPP-38.901-Indoor-LOS}&  Mean  &  -0.104 &  0.104 &  0.106 &  -0.122 &  -0.052\\
 &  &  Std. & 0.358 & 0.39 & 0.404 & 0.367 & 0.322\\

 & \Block{2-1}{mmMAGIC-Indoor-LOS}&  Mean  &  0.037 &  -0.011 &  -0.018 &  0.016 &  -0.046\\
 &  &  Std. & 0.174 & 0.153 & 0.154 & 0.16 & 0.193\\
\Xhline{4\arrayrulewidth}

\Block{6-1}{V103}&\Block{2-1}{QuaDRiGa-Industrial-LOS}&  Mean  &  0.043 &  -0.065 &  1.232 &  -0.066 &  -0.387\\
 &  &  Std. & 0.775 & 0.784 & 1.628 & 0.772 & 1.275\\

 & \Block{2-1}{3GPP-38.901-Indoor-LOS}&  Mean  &  -0.042 &  0.053 &  0.008 &  -0.033 &  -0.022\\
 &  &  Std. & 0.353 & 0.382 & 0.387 & 0.36 & 0.369\\

 & \Block{2-1}{mmMAGIC-Indoor-LOS}&  Mean  &  0.001 &  -0.011 &  0.012 &  0.0 &  -0.013\\
 &  &  Std. & 0.176 & 0.166 & 0.17 & 0.18 & 0.183\\
\Xhline{4\arrayrulewidth}

\Block{6-1}{V201}&\Block{2-1}{QuaDRiGa-Industrial-LOS}&  Mean  &  0.059 &  0.108 &  -0.182 &  -0.154 &  -0.27\\
 &  &  Std. & 0.751 & 0.897 & 0.592 & 0.986 & 0.79\\

 & \Block{2-1}{3GPP-38.901-Indoor-LOS}&  Mean  &  0.025 &  -0.07 &  -0.045 &  0.054 &  0.12\\
 &  &  Std. & 0.364 & 0.379 & 0.392 & 0.302 & 0.367\\

 & \Block{2-1}{mmMAGIC-Indoor-LOS}&  Mean  &  -0.012 &  0.026 &  0.015 &  -0.019 &  0.015\\
 &  &  Std. & 0.164 & 0.177 & 0.163 & 0.18 & 0.192\\
\Xhline{4\arrayrulewidth}

\Block{6-1}{V202}&\Block{2-1}{QuaDRiGa-Industrial-LOS}&  Mean  &  0.027 &  0.141 &  0.072 &  0.082 &  -0.204\\
 &  &  Std. & 0.716 & 0.674 & 0.908 & 0.933 & 0.631\\

 & \Block{2-1}{3GPP-38.901-Indoor-LOS}&  Mean  &  0.052 &  -0.053 &  -0.039 &  0.012 &  0.043\\
 &  &  Std. & 0.391 & 0.348 & 0.427 & 0.359 & 0.351\\

 & \Block{2-1}{mmMAGIC-Indoor-LOS}&  Mean  &  -0.007 &  0.007 &  -0.018 &  0.013 &  0.011\\
 &  &  Std. & 0.178 & 0.168 & 0.17 & 0.19 & 0.192\\
\Xhline{4\arrayrulewidth}

\Block{6-1}{V203}&\Block{2-1}{QuaDRiGa-Industrial-LOS}&  Mean  &  0.067 &  -0.017 &  0.273 &  -0.13 &  -0.045\\
 &  &  Std. & 0.754 & 0.73 & 1.343 & 0.724 & 0.684\\

 & \Block{2-1}{3GPP-38.901-Indoor-LOS}&  Mean  &  0.009 &  -0.022 &  -0.046 &  0.043 &  0.02\\
 &  &  Std. & 0.376 & 0.351 & 0.317 & 0.37 & 0.358\\

 & \Block{2-1}{mmMAGIC-Indoor-LOS}&  Mean  &  0.018 &  0.017 &  0.002 &  0.004 &  -0.009\\
 &  &  Std. & 0.17 & 0.181 & 0.174 & 0.178 & 0.174\\
\Xhline{4\arrayrulewidth}

\end{NiceTabular}
}
\end{table*}

\subsection{Evaluation Metrics}

For the evaluation of our approach, we utilize the two most popular metrics in Simultaneous Localization and Mapping (SLAM): Absolute Trajectory Error (ATE) and Relative Pose Error (RPE)~\cite{prokhorov2019measuring}. ATE is a metric for global consistency, defined as the Root Mean Square Error (RMSE) between the absolute distances of the estimated and ground truth trajectories. RPE complements ATE by focusing on local accuracy and is primarily utilized for evaluating odometry systems. It provides a concise measure of the error between consecutive poses in the estimated trajectory compared to the ground truth trajectory. RPE helps to assess the drift or error accumulation in pose estimation over time, specifically at smaller time intervals.

In addition to these metrics, we analyzed the RMSE for each trajectory coordinate axis, denoted as $E_{a}$ where $a$ represents $x$, $y$, or $z$. This calculation was carried out to assess the accuracy of the estimation along specific spatial directions. The $\RMSE$ error for each coordinate axis $a \in \{ x, y, z\}$ was calculated using the following formula:

\begin{equation*}
E_{a} = \sqrt{\frac{1}{N}\sum_{i=1}^{N}(a_{i,\text{est}} - a_{i,\text{gt}})^2}, \quad a \in \{ x, y, z\}
\end{equation*}
In this equation, $N$ represents the total number of poses, while the subscripts $(i, \text{est})$ and $(i, \text{gt})$ correspond to the $i$-th estimated and ground truth poses, respectively. This analysis gave us valuable insights into potential accuracy variations dependent on the spatial direction. By evaluating the error along each axis separately, we could identify any discrepancies in performance for different spatial orientations.

To further assess the suitability of the proposed algorithms for real-time applications, we evaluated their implementation times. For PGO, we calculated the average and standard deviation (std) of the time taken for each optimization process. For the ESKF, we recorded the average and std of the time taken to complete one prediction plus update state cycle.

\subsection{Results}
\begin{table*}
\setlength\tabcolsep{3.3pt}
\renewcommand{\arraystretch}{1.2}
\centering\footnotesize
    \caption{Evaluation of the Error State Kalman Filter (ESKF) vs. Graph-Based Optimization Localization under Different Scenarios and Configurations for Vicon Room 1. \textbf{Best values for each dataset are shown in bold.}}
    \label{tab:VR1}
\resizebox{0.8\linewidth}{!}{
\begin{NiceTabular}{M{1.5cm}M{2cm}M{1.5cm}M{1.5cm}M{1.5cm}M{1.5cm}M{1.5cm}M{1.5cm}M{1.5cm}P{1.7cm}M{1.5cm}}[hvlines]
\CodeBefore
  \rectanglecolor{mygray}{2-3}{5-11}
  \rectanglecolor{mygray}{10-3}{13-11}
  \rectanglecolor{mygray}{18-3}{21-11}
  \rectanglecolor{mygray}{26-3}{29-11}
  \rectanglecolor{mygray}{34-3}{37-11}
  \rectanglecolor{mygray}{42-3}{45-11}
  \rectanglecolor{mygray}{50-3}{53-11}
  \rectanglecolor{mygray}{58-3}{61-11}
  \rectanglecolor{mygray}{66-3}{69-11}
  \rectanglecolor{mygray}{74-3}{77-11}
\Body
    \Xhline{4\arrayrulewidth}
    \bf Dataset            & \bf 5G sim. scenario & \bf Approach                    & \bf BS Num & \bf ATE(m)   & \bf  $\mathbf{E}_{\mathbf{X}}$(m) & \bf $\mathbf{E}_{\mathbf{Y}}$(m) & \bf $\mathbf{E}_{\mathbf{Z}}$(m) & \bf $\mathbf{RPE}_\mathbf{T}$ (m) & \bf $\mathbf{RPE}_\mathbf{R}$ (deg)&  \bf $\mathbf{ATE}_\mathbf{avg.}$ \\

\Block{24-1}{V101}&\Block{8-1}{QuaDRiGa-Industrial-LOS} & \Block{4-1}{Graph-based} & 2 & 3.7290 & 0.7326 & 1.1775 & 3.4615 & 0.0174 & 0.3979 & \Block{4-1}{1.438}  \\
 &  &  & 3 & 1.2566 & 0.2978 & 0.1910 & 1.2058 & 0.0113 & 0.3984 &  \\
 &  &  & 4 & 1.5274 & 0.2230 & 0.1588 & 1.5026 & 0.0091 & 0.4175 &  \\
 &  &  & 5 & 0.6791 & 0.2431 & 0.1553 & 0.6147 & 0.0082 & 0.4142 &  \\

& & \Block{4-1}{ESKF-based} & 2 & 6.6928 & 3.2562 & 4.3922 & 3.8600 & 0.5830 & 0.0286 & \Block{4-1}{3.17}  \\
& &  & 3 & 2.9209 & 0.7723 & 0.6821 & 2.7331 & 0.2571 & 0.0211 &  \\
& &  & 4 & 3.3593 & 0.6545 & 0.5430 & 3.2498 & 0.2672 & 0.0244 &  \\
& &  & 5 & 2.8782 & 0.5868 & 0.4830 & 2.7760 & 0.2644 & \bf 0.0201 &  \\

 & \Block{8-1}{3GPP-38.901-Indoor-LOS} & \Block{4-1}{Graph-based} & 2 & 2.5805 & 1.0242 & 1.6213 & 1.7266 & 0.0182 & 0.3979 & \Block{4-1}{0.867}  \\
 &  &  & 3 & 1.1189 & 0.1935 & 0.1225 & 1.0952 & 0.0083 & 0.3979 &  \\
 &  &  & 4 & 0.3787 & 0.1757 & 0.1147 & 0.3152 & 0.0070 & 0.4114 &  \\
 &  &  & 5 & 0.2583 & 0.1717 & 0.0970 & 0.1668 & 0.0066 & 0.4487 &  \\

& & \Block{4-1}{ESKF-based} & 2 & 5.1227 & 2.5365 & 2.8306 & 3.4345 & 0.3837 & 0.0333 & \Block{4-1}{1.934}  \\
& &  & 3 & 2.2327 & 0.5324 & 0.3538 & 2.1392 & 0.2465 & 0.0209 &  \\
& &  & 4 & 1.4097 & 0.4463 & 0.2755 & 1.3084 & 0.1865 & 0.0356 &  \\
& &  & 5 & 0.9072 & 0.4194 & 0.2373 & 0.7687 & 0.1763 & 0.0266 &  \\

 & \Block{8-1}{mmMAGIC-Indoor-LOS} & \Block{4-1}{Graph-based} & 2 & 1.9861 & 0.7814 & 1.0087 & 1.5220 & 0.0104 &  0.3978 & \Block{4-1}{\bf 0.701}  \\
 &  &  & 3 & 1.2447 & 0.1228 & 0.0522 & 1.2375 & 0.0065 & 0.4378 &  \\
 &  &  & 4 & 0.1432 & 0.0576 & 0.0468 & 0.1225 & 0.0055 & 0.4468 &  \\
 &  &  & 5 & \bf 0.1312 & \bf 0.0516 & \bf 0.0422 & \bf 0.1131 & \bf 0.0052 & 0.4571 &  \\

& & \Block{4-1}{ESKF-based} & 2 & 5.0005 & 2.2481 & 3.3849 & 2.9143 & 0.4272 & 0.0255 & \Block{4-1}{1.504}  \\
& &  & 3 & 1.7167 & 0.2226 & 0.1824 & 1.6924 & 0.2070 & 0.0337 &  \\
& &  & 4 & 0.4643 & 0.1745 & 0.1481 & 0.4040 & 0.1596 & 0.0472 &  \\
& &  & 5 & 0.3400 & 0.1602 & 0.1305 & 0.2699 & 0.1436 & 0.0475 &  \\

\Xhline{4\arrayrulewidth}
\Block{24-1}{V102}&\Block{8-1}{QuaDRiGa-Industrial-LOS} & \Block{4-1}{Graph-based} & 2 & 1.5352 & 0.8783 & 0.4125 & 1.1897 & 0.0110 & 0.4361 & \Block{4-1}{0.744}  \\
 &  &  & 3 & 0.8399 & 0.2819 & 0.1810 & 0.7702 & 0.0108 & 0.4373 &  \\
 &  &  & 4 & 0.7252 & 0.2178 & 0.1785 & 0.6683 & 0.0102 & 0.4365 &  \\
 &  &  & 5 & 0.6216 & 0.2006 & 0.1540 & 0.5678 & 0.0094 & 0.4373 &  \\

& & \Block{4-1}{ESKF-based} & 2 & 4.6070 & 3.4935 & 0.9800 & 2.8389 & 0.4711 & 0.0394 & \Block{4-1}{2.154}  \\
& &  & 3 & 2.1624 & 0.9664 & 0.7710 & 1.7742 & 0.3553 & 0.0672 &  \\
& &  & 4 & 2.0928 & 0.8675 & 0.7558 & 1.7482 & 0.3709 & 0.1039 &  \\
& &  & 5 & 1.9070 & 0.7949 & 0.7851 & 1.5455 & 0.3664 & 0.1051 &  \\

 & \Block{8-1}{3GPP-38.901-Indoor-LOS} & \Block{4-1}{Graph-based} & 2 & 0.7864 & 0.4475 & 0.1228 & 0.6349 & 0.0094 & 0.4361 & \Block{4-1}{0.523}  \\
 &  &  & 3 & 1.3843 & 0.1719 & 0.1388 & 1.3666 & 0.0123 & 0.4363 &  \\
 &  &  & 4 & 0.2257 & 0.1620 & 0.0983 & 0.1226 & 0.0083 & 0.4381 &  \\
 &  &  & 5 & 0.2171 & 0.1588 & 0.0946 & 0.1139 & 0.0082 & 0.4449 &  \\

& & \Block{4-1}{ESKF-based} & 2 & 5.0163 & 4.3755 & 1.2167 & 2.1302 & 0.5266 & \bf 0.0345 & \Block{4-1}{1.625}  \\
& &  & 3 & 1.2971 & 0.6606 & 0.4610 & 1.0166 & 0.3490 & 0.0819 &  \\
& &  & 4 & 0.9720 & 0.4812 & 0.5191 & 0.6661 & 0.3181 & 0.0796 &  \\
& &  & 5 & 0.8386 & 0.4801 & 0.4566 & 0.5142 & 0.2920 & 0.0794 &  \\

 & \Block{8-1}{mmMAGIC-Indoor-LOS} & \Block{4-1}{Graph-based} & 2 & 0.3714 & 0.1816 & 0.0661 & 0.3172 & 0.0084 & 0.4362 & \Block{4-1}{\bf 0.166}  \\
 &  &  & 3 & 0.1981 & 0.0640 & 0.0451 & 0.1820 & 0.0074 & 0.4418 &  \\
 &  &  & 4 & 0.1297 & \bf 0.0527 & 0.0441 & 0.1100 & 0.0074 & 0.4653 &  \\
 &  &  & 5 & \bf 0.1286 & 0.0603 & \bf 0.0365 & \bf 0.1076 & \bf 0.0072 & 0.4549 &  \\

& & \Block{4-1}{ESKF-based} & 2 & 4.3145 & 3.3932 & 0.5532 & 2.6068 & 0.5891 & 0.0454 & \Block{4-1}{1.409}  \\
& &  & 3 & 1.7085 & 0.3224 & 0.3215 & 1.6467 & 0.3436 & 0.0482 &  \\
& &  & 4 & 0.5381 & 0.2799 & 0.3225 & 0.3275 & 0.2720 & 0.1526 &  \\
& &  & 5 & 0.4848 & 0.2804 & 0.2686 & 0.2902 & 0.2544 & 0.1653 &  \\

\Xhline{4\arrayrulewidth}
\Block{24-1}{V103}&\Block{8-1}{QuaDRiGa-Industrial-LOS} & \Block{4-1}{Graph-based} & 2 & 1.7425 & 0.8096 & 0.4779 & 1.4671 & 0.0185 & 0.7368 & \Block{4-1}{1.247}  \\
 &  &  & 3 & 2.1882 & 0.4163 & 0.2007 & 2.1388 & 0.0141 & 0.7372 &  \\
 &  &  & 4 & 0.9904 & 0.3355 & 0.1479 & 0.9200 & 0.0156 & 0.7381 &  \\
 &  &  & 5 & 1.3146 & 0.6974 & 0.3194 & 1.0676 & 0.0173 & 0.7443 &  \\

& & \Block{4-1}{ESKF-based} & 2 & 4.7003 & 3.7447 & 1.2424 & 2.5547 & 0.4716 & 0.0373 & \Block{4-1}{3.237}  \\
& &  & 3 & 4.4011 & 0.8701 & 0.9770 & 4.2022 & 0.3801 & 0.0397 &  \\
& &  & 4 & 3.2805 & 0.6970 & 0.8300 & 3.0963 & 0.3068 & 0.0657 &  \\
& &  & 5 & 3.8017 & 1.1043 & 0.9159 & 3.5206 & 0.3981 & 0.0690 &  \\

 & \Block{8-1}{3GPP-38.901-Indoor-LOS} & \Block{4-1}{Graph-based} & 2 & 0.7494 & 0.4310 & 0.3216 & 0.5219 & 0.0133 & 0.7363 & \Block{4-1}{\bf 0.355}  \\
 &  &  & 3 & 0.4769 & 0.2144 & 0.0727 & 0.4198 & 0.0117 & 0.7412 &  \\
 &  &  & 4 & 0.2759 & 0.2287 & 0.0685 & 0.1383 & 0.0113 & 0.7446 &  \\
 &  &  & 5 & 0.2733 & 0.2242 & 0.0538 & 0.1467 & 0.0113 & 0.7492 &  \\

& & \Block{4-1}{ESKF-based} & 2 & 4.8341 & 3.9437 & 1.2121 & 2.5193 & 0.3909 & \bf 0.0355 & \Block{4-1}{2.028}  \\
& &  & 3 & 2.4700 & 0.8539 & 0.6351 & 2.2290 & 0.2908 & 0.0964 &  \\
& &  & 4 & 1.6018 & 0.8075 & 0.5747 & 1.2584 & 0.2673 & 0.1084 &  \\
& &  & 5 & 1.2320 & 0.8089 & 0.4765 & 0.7978 & 0.2530 & 0.0887 &  \\

 & \Block{8-1}{mmMAGIC-Indoor-LOS} & \Block{4-1}{Graph-based} & 2 & 1.2840 & 0.2598 & 0.1496 & 1.2485 & 0.0112 & 0.7362 & \Block{4-1}{0.558}  \\
 &  &  & 3 & 1.2075 & 0.0970 & 0.0603 & 1.2021 & 0.0120 & 0.7453 &  \\
 &  &  & 4 & \bf 0.1365 & \bf 0.0738 & 0.0510 & \bf 0.1029 & 0.0105 & 0.7720 &  \\
 &  &  & 5 & 0.1614 & 0.0754 & \bf 0.0398 &  0.1370 & \bf 0.0102 & 0.7636 &  \\

& & \Block{4-1}{ESKF-based} & 2 & 3.8474 & 2.9339 & 1.3210 & 2.1094 & 0.4834 & 0.0596 & \Block{4-1}{1.388}  \\
& &  & 3 & 1.4482 & 0.6386 & 0.3389 & 1.2548 & 0.2525 & 0.1338 &  \\
& &  & 4 & 0.7638 & 0.5816 & 0.3572 & 0.3426 & 0.2104 & 0.1402 &  \\
& &  & 5 & 0.8789 & 0.6446 & 0.3093 & 0.5111 & 0.2092 & 0.1180 &  \\

\Xhline{4\arrayrulewidth}
\end{NiceTabular}
}
\end{table*}

\begin{table*}
\setlength\tabcolsep{3.3pt}
\renewcommand{\arraystretch}{1.2}
\centering\footnotesize
    \caption{Evaluation of the Error State Kalman Filter (ESKF) vs. Graph-Based Optimization Localization under Different Scenarios and Configurations for Vicon Room 2. \textbf{Best values for each dataset are shown in bold.}}
    \label{tab:VR2}
\resizebox{0.8\linewidth}{!}{
\begin{NiceTabular}{M{1.5cm}M{2cm}M{1.5cm}M{1.5cm}M{1.5cm}M{1.5cm}M{1.5cm}M{1.5cm}M{1.5cm}P{1.7cm}M{1.5cm}}[hvlines]
\CodeBefore
  \rectanglecolor{mygray}{2-3}{5-11}
  \rectanglecolor{mygray}{10-3}{13-11}
  \rectanglecolor{mygray}{18-3}{21-11}
  \rectanglecolor{mygray}{26-3}{29-11}
  \rectanglecolor{mygray}{34-3}{37-11}
  \rectanglecolor{mygray}{42-3}{45-11}
  \rectanglecolor{mygray}{50-3}{53-11}
  \rectanglecolor{mygray}{58-3}{61-11}
  \rectanglecolor{mygray}{66-3}{69-11}
  \rectanglecolor{mygray}{74-3}{77-11}
\Body
    \Xhline{4\arrayrulewidth}
    \bf Dataset            & \bf 5G sim. scenario & \bf Approach                    & \bf BS Num & \bf ATE(m)   & \bf  $\mathbf{E}_{\mathbf{X}}$(m) & \bf $\mathbf{E}_{\mathbf{Y}}$(m) & \bf $\mathbf{E}_{\mathbf{Z}}$(m) & \bf $\mathbf{RPE}_\mathbf{T}$ (m) & \bf $\mathbf{RPE}_\mathbf{R}$ (deg)& \bf $\mathbf{ATE}_\mathbf{avg.}$ \\
\Block{24-1}{V201}&\Block{8-1}{QuaDRiGa-Industrial-LOS} & \Block{4-1}{Graph-based} & 2 & 5.2234 & 3.7356 & 2.7495 & 2.4020 & 0.0399 & 1.2588 & \Block{4-1}{2.319}  \\
 &  &  & 3 & 2.6128 & 0.5229 & 0.5995 & 2.4887 & 0.0246 & 1.2620 &  \\
 &  &  & 4 & 2.0649 & 0.4159 & 0.3468 & 1.9926 & 0.0204 & 1.2601 &  \\
 &  &  & 5 & 1.6947 & 0.4047 & 0.2766 & 1.6223 & 0.0188 & 1.2640 &  \\

& & \Block{4-1}{ESKF-based} & 2 & 5.1291 & 3.1488 & 2.4413 & 3.2299 & 0.4138 & \bf 0.0454 & \Block{4-1}{2.939}  \\
& &  & 3 & 3.9739 & 0.7953 & 0.9333 & 3.7800 & 0.3330 & 0.0483 &  \\
& &  & 4 & 2.6749 & 0.5321 & 0.6420 & 2.5416 & 0.2257 & 0.0476 &  \\
& &  & 5 & 2.9173 & 0.6072 & 0.4912 & 2.8108 & 0.2127 & 0.0800 &  \\

 & \Block{8-1}{3GPP-38.901-Indoor-LOS} & \Block{4-1}{Graph-based} & 2 & 2.3833 & 1.5583 & 1.4689 & 1.0461 & 0.0239 & 1.2580 & \Block{4-1}{\bf 0.875}  \\
 &  &  & 3 & 1.2636 & 0.2385 & 0.2003 & 1.2246 & 0.0153 & 1.2581 &  \\
 &  &  & 4 & 0.3906 & 0.1807 & 0.1700 & 0.3017 & 0.0133 & 1.2771 &  \\
 &  &  & 5 & 0.3375 & 0.1792 & 0.1300 & 0.2548 & 0.0130 & 1.2656 &  \\

& & \Block{4-1}{ESKF-based} & 2 & 4.1948 & 2.1934 & 1.5755 & 3.2099 & 0.4360 & 0.0623 & \Block{4-1}{1.68}  \\
& &  & 3 & 2.1001 & 0.5156 & 0.3570 & 2.0043 & 0.2055 & 0.0501 &  \\
& &  & 4 & 1.2208 & 0.3549 & 0.3474 & 1.1152 & 0.1684 & 0.0495 &  \\
& &  & 5 & 0.8838 & 0.3682 & 0.2477 & 0.7644 & 0.1535 & 0.0528 &  \\

 & \Block{8-1}{mmMAGIC-Indoor-LOS} & \Block{4-1}{Graph-based} & 2 & 4.1182 & 2.6694 & 2.0691 & 2.3564 & 0.0355 & 1.2578 & \Block{4-1}{1.142}  \\
 &  &  & 3 & 1.1002 & 1.1678 & 0.0602 & 1.0856 & 0.0158 & 1.2626 &  \\
 &  &  & 4 & 0.2598 & \bf 0.0783 & 0.0549 & 0.2416 & 0.0128 & 1.2730 &  \\
 &  &  & 5 & \bf 0.2307 & 0.0797 & \bf 0.0475 & \bf 0.2113 & \bf 0.0123 & 1.2633 &  \\

& & \Block{4-1}{ESKF-based} & 2 & 3.9157 & 2.1876 & 2.2021 & 2.3870 & 0.3262 & 0.0907 & \Block{4-1}{1.371}  \\
& &  & 3 & 1.8596 & 0.2765 & 0.2222 & 1.8254 & 0.1780 & 0.0509 &  \\
& &  & 4 & 0.6011 & 0.2386 & 0.1851 & 0.5197 & 0.1489 & 0.0554 &  \\
& &  & 5 & 0.4789 & 0.2505 & 0.1974 & 0.3573 & 0.1259 & 0.0577 &  \\

\Xhline{4\arrayrulewidth}
\Block{24-1}{V202}&\Block{8-1}{QuaDRiGa-Industrial-LOS} & \Block{4-1}{Graph-based} & 2 & 2.6935 & 1.5963 & 1.2222 & 1.7925 & 0.0367 & 2.7873 & \Block{4-1}{1.453}  \\
 &  &  & 3 & 2.4520 & 0.6049 & 0.4874 & 2.3256 & 0.0321 & 2.7877 &  \\
 &  &  & 4 & 1.4682 & 0.3855 & 0.2990 & 1.3847 & 0.0300 & 2.7900 &  \\
 &  &  & 5 & 0.6496 & 0.3859 & 0.2206 & 0.4737 & 0.0283 & 2.7902 &  \\

& & \Block{4-1}{ESKF-based} & 2 & 5.6998 & 3.5542 & 3.1184 & 3.1830 & 0.6514 & 0.1137 & \Block{4-1}{3.038}  \\
& &  & 3 & 3.4033 & 1.2179 & 0.9019 & 3.0472 & 0.4557 & 0.0987 &  \\
& &  & 4 & 3.3419 & 0.7543 & 0.5956 & 3.2007 & 0.3051 & 0.0986 &  \\
& &  & 5 & 2.7431 & 0.8620 & 0.5499 & 2.5454 & 0.3003 & 0.1049 &  \\

 & \Block{8-1}{3GPP-38.901-Indoor-LOS} & \Block{4-1}{Graph-based} & 2 & 1.7066 & 0.7871 & 0.7965 & 1.2878 & 0.0348 & 2.7873 & \Block{4-1}{0.738}  \\
 &  &  & 3 & 1.0422 & 0.2209 & 0.1824 & 1.0020 & 0.0294 & 2.7881 &  \\
 &  &  & 4 & 0.5193 & 0.1608 & 0.1479 & 0.4712 & 0.0277 & 2.7915 &  \\
 &  &  & 5 & 0.4228 & 0.1740 & 0.1319 & 0.3621 & 0.0274 & 2.7903 &  \\

& & \Block{4-1}{ESKF-based} & 2 & 3.9675 & 2.3513 & 1.9427 & 2.5373 & 0.4971 & \bf 0.0962 & \Block{4-1}{1.634}  \\
& &  & 3 & 2.1767 & 0.5870 & 0.4274 & 2.0520 & 0.3171 & 0.1224 &  \\
& &  & 4 & 1.0396 & 0.4805 & 0.3843 & 0.8380 & 0.2640 & 0.1300 &  \\
& &  & 5 & 0.9861 & 0.5676 & 0.3499 & 0.7264 & 0.2564 & 0.1245 &  \\

 & \Block{8-1}{mmMAGIC-Indoor-LOS} & \Block{4-1}{Graph-based} & 2 & 1.4226 & 0.7650 & 0.6926 & 0.9792 & 0.0317 & 2.7873 & \Block{4-1}{\bf 0.559}  \\
 &  &  & 3 & 0.9065 & 0.1162 & 0.1054 & 0.8929 & 0.0282 & 2.7887 &  \\
 &  &  & 4 & 0.2417 & \bf 0.1000 & 0.0961 & 0.1980 & 0.02682 & 2.7960 &  \\
 &  &  & 5 & \bf 0.2244 & 0.1012 & \bf 0.0909 & \bf 0.1785 & \bf 0.02680 & 2.8014 &  \\

& & \Block{4-1}{ESKF-based} & 2 & 4.5127 & 3.1000 & 2.6701 & 1.9040 & 0.4324 & 0.1106 & \Block{4-1}{1.462}  \\
& &  & 3 & 1.4249 & 0.4480 & 0.3224 & 1.3136 & 0.2629 & 0.2437 &  \\
& &  & 4 & 0.7655 & 0.4097 & 0.2682 & 0.5884 & 0.2236 & 0.2483 &  \\
& &  & 5 & 0.6050 & 0.2987 & 0.2638 & 0.4551 & 0.2163 & 0.1951 &  \\

\Xhline{4\arrayrulewidth}
\Block{24-1}{V203}&\Block{8-1}{QuaDRiGa-Industrial-LOS} & \Block{4-1}{Graph-based} & 2 & 3.4606 & 2.3413 & 2.2257 & 1.2410 & 0.0442 & 3.2686 & \Block{4-1}{1.258}  \\
 &  &  & 3 & 0.9769 & 0.3589 & 0.3384 & 0.8432 & 0.0274 & 3.2689 &  \\
 &  &  & 4 & 1.2247 & 0.2967 & 0.3169 & 1.1452 & 0.0295 & 3.2717 &  \\
 &  &  & 5 & 0.6272 & 0.3255 & 0.2812 & 0.4565 & 0.0269 & 3.2758 &  \\

& & \Block{4-1}{ESKF-based} & 2 & 5.5247 & 2.9753 & 2.8260 & 3.6992 & 0.6252 & \bf 0.1131 & \Block{4-1}{2.758}  \\
& &  & 3 & 3.0756 & 0.9551 & 0.9707 & 2.7577 & 0.3711 & 0.1402 &  \\
& &  & 4 & 3.0525 & 0.8027 & 0.6954 & 2.8618 & 0.3478 & 0.1223 &  \\
& &  & 5 & 2.1353 & 0.9533 & 0.6151 & 1.8090 & 0.3513 & 0.1257 &  \\

 & \Block{8-1}{3GPP-38.901-Indoor-LOS} & \Block{4-1}{Graph-based} & 2 & 3.0104 & 1.8298 & 1.8757 & 1.4819 & 0.0342 & 3.2686 & \Block{4-1}{0.962}  \\
 &  &  & 3 & 0.9943 & 0.1643 & 0.1753 & 0.9648 & 0.0265 & 3.2686 &  \\
 &  &  & 4 & 0.4270 & 0.1577 & 0.1328 & 0.3740 & 0.0258 & 3.2754 &  \\
 &  &  & 5 & 0.3771 & 0.1526 & 0.1100 & 0.3268 & 0.0258 & 3.2775 &  \\

& & \Block{4-1}{ESKF-based} & 2 & 4.0704 & 2.2813 & 1.9862 & 2.7238 & 0.4975 & 0.1227 & \Block{4-1}{1.568}  \\
& &  & 3 & 1.7312 & 0.5890 & 0.4748 & 1.5571 & 0.2855 & 0.1418 &  \\
& &  & 4 & 1.0466 & 0.4893 & 0.4362 & 0.8159 & 0.2687 & 0.1405 &  \\
& &  & 5 & 0.9896 & 0.4989 & 0.4095 & 0.7501 & 0.2498 & 0.1412 &  \\

 & \Block{8-1}{mmMAGIC-Indoor-LOS} & \Block{4-1}{Graph-based} & 2 & 1.5031 & 0.5633 & 0.6366 & 1.2397 & 0.0290 & 3.2686 & \Block{4-1}{\bf 0.475}  \\
 &  &  & 3 & 0.4342 & 0.1250 & 0.0858 & 0.4068 & 0.0263 & 3.2697 &  \\
 &  &  & 4 & 0.2509 & \bf 0.1202 & 0.0855 & 0.2030 & 0.0257 & 3.2752 &  \\
 &  &  & 5 & \bf 0.1878 & 0.1224 & \bf 0.0782 & \bf 0.1191 & \bf 0.0254 & 3.2771 &  \\

& & \Block{4-1}{ESKF-based} & 2 & 4.2958 & 2.2371 & 2.5190 & 2.6654 & 0.5333 & 0.1149 & \Block{4-1}{1.392}  \\
& &  & 3 & 1.6151 & 0.2677 & 0.3013 & 1.5640 & 0.2796 & 0.1407 &  \\
& &  & 4 & 0.6045 & 0.2814 & 0.2341 & 0.4811 & 0.2218 & 0.1482 &  \\
& &  & 5 & 0.4459 & 0.2554 & 0.2195 & 0.2923 & 0.1992 & 0.1569 &  \\

\Xhline{4\arrayrulewidth}

\end{NiceTabular}
}
\end{table*}

The experiments were performed on a Ubuntu 20.04 laptop with an Intel(R) Core(TM) i9-10885H CPU @ 2.40GHz with 16 cores and 32 Gb of RAM. All codes (PGO and ESKF) are implemented in Python, utilizing the relative interface of GTSAM v4.0 for creating and optimizing the factor graphs.
The drone's position and orientation results are obtained from the factor graph based on the final Maximum A Posteriori (MAP) estimate for each node, where nodes are consistently generated at 10 Hz, twice the frequency of \toa measurements. On the other hand, the results for the Error State Kalman Filter (ESKF) approach are acquired at each update stage, coinciding with the reception of \toa measurements. To comprehensively evaluate the performance of the localization algorithm, error metrics are computed by establishing a comparison between the ground truth Vicon pose and the estimated pose that is temporally closest.

The detailed results for the graph-based and ESKF methods are exhaustively documented in \autoref{tab:VR1} and \autoref{tab:VR2}, corresponding to Vicon Room 1 and Vicon Room 2, respectively. These tables provide comprehensive information encompassing ATE, RPE, and translation RMSE for each distinct motion direction in various 5G simulation scenarios for both methods.  Additionally, the ATE\textsubscript{avg.} column presents the average ATE for each method across each scenario with 2, 3, 4, and 5 base stations.

Subsequently, we delve into a detailed analysis of the obtained results, uncovering the underlying factors contributing to the observed performance trends. 

\begin{itemize}
        \item \textbf{Base Stations and Bandwidth Influence:} The results consistently show that the accuracy of \toa-based localization improves when the number of base stations and the available bandwidth (communication scenarios) increase. This is because more base stations provide more reference points for \toa measurements, and higher bandwidth allows for higher resolution \toa measurements. For example, using dataset V101 and a fixed communication scenario (3GPP-38.901-Indoor-LOS), the error decreases from $2.58~m$ to $0.25~m$ under graph-based optimization when the number of base stations increases from 2 to 5. However, adding more than four base stations sometimes does not lead to significant or further improvement. This may indicate a lower bound to the error reduction achieved by adding more antennas. Nevertheless, such redundancy may be helpful in those environments where NLOS conditions are more frequent. The effect of bandwidth on the accuracy is also evident, as higher bandwidth leads to higher resolution \toa measurement. In particular, the third scenario (mmMAGIC-Indoor-LOS), which has the highest bandwidth (400 MHz), also offers the highest accuracy for the same antennas.

        \item \textbf{Reduced Accuracy in the Vertical Dimension (z-Axis)} Upon analyzing the Root Mean Square Error (RMSE) along each axis ($E_x$, $E_y$, $E_z$), a prominent observation is the relatively larger errors in the $E_z$ component, indicative of the vertical direction estimation challenge. This is attributed to the limited offsets provided by base stations in the vertical direction compared to the $x$ and $y$ directions.
    
        \item \textbf{Limited Impact of \toa Measurements on Attitude Estimation:} In assessing attitude estimation, a lack of discernible patterns is apparent, suggesting that the inclusion of \toa measurements does not notably enhance the accuracy of attitude estimation. This is inherently expected as \toa measurements primarily offer information about position rather than attitude. 
    
        \item \textbf{Superior Performance of Graph-Based Optimization Compared to ESKF:} Notably, the graph-based optimization consistently outperforms ESKF across all datasets, scenarios, and base station configurations, demonstrating the graph-based approach's superior efficacy in enhancing indoor localization accuracy. The main reason behind this advantage lies in the filtering method's practice of marginalizing all older information by limiting itself to the most recent states. In contrast, the graph-based method utilizes the entire historical information, employing all past measurements up to the current one and optimizing the entire trajectory. Notably, graph-based methods tend to perform better when dealing with sparse, low-frequency measurements, such as our implementation utilizing \toa measurements with a frequency of 5 Hz, which is relatively low. This increased susceptibility of the ESKF to sparse measurements can contribute to more significant performance disparities between the two methods.

        \item \textbf{Robustness Across Different Trajectories and Datasets:} We assessed the robustness of our proposed approaches across datasets with varying levels of complexity, encompassing sequences from two different Vicon rooms: Vicon Room 1 (V101, V102, V103) and Vicon Room 2 (V201, V202, V203). These sequences were categorized into three levels of difficulty: easy (V101, V201), medium (V102, V202), and difficult (V103, V203).

        Despite the diversity in trajectory complexity and environmental conditions, we observed that the accuracy of our proposed approaches did not exhibit a clear pattern across the different sequences. Surprisingly, in some instances, our methods yielded better results for the more difficult datasets, challenging conventional expectations. This variability underscores the importance of robustness and adaptability in MAV pose estimation algorithms, especially when operating in diverse and dynamic environments.
        
        Our findings highlight the effectiveness of our proposed methods in handling the complexities inherent in MAV flight scenarios. The lack of a consistent pattern in accuracy underscores the versatility of our approaches, demonstrating their ability to adapt to various challenges encountered during indoor flight missions. This adaptability is crucial for real-world applications, where MAVs must navigate through unpredictable environments with confidence and precision.

\end{itemize}
{Table~\ref{tab:execution_time} presents the execution times for both PGO and ESKF, showcasing the best results achieved using five base stations. As evident from the table, both approaches are suitable for real-time implementation, with ESKF outperforming PGO, as expected. This observation reinforces ESKF's superior efficiency for real-time applications.
For instance, in the V101 dataset, PGO required an average of 3.08 milliseconds per optimization, while ESKF consumed only 0.39 milliseconds for a prediction-update cycle. Since the codes were implemented in Python (with PGO partially implemented in Python), we expect further improvement in real-time performance by converting the codes entirely to C++. Overall, while the ESKF estimation in general runs faster than the PGO, these results also show that both estimation approaches can be implemented for real-time MAV applications, given a suitable on-board processing unit.

\begin{table*}
    \centering
    \caption{Computation time comparison for PGO and ESKF in Euroc MAV dataset localization. The table showcases the best results for each dataset, including the average time and standard deviation for each PGO optimization and ESKF (both prediction and update steps). All times are reported in milliseconds.}
    \label{tab:execution_time} 
    \resizebox{0.8\linewidth}{!}{
        \begin{tabular}{ccccccc}
        \Xhline{3\arrayrulewidth}
        & \multicolumn{6}{c}{\bf Dataset} \\
        \cmidrule(lr){2-7}
        
            \bf Method  & \bf V101                    & \bf V102 & \bf V103   & \bf  V201 & \bf V202 & \bf V203  \\

    \cmidrule(lr){1-1} \cmidrule(lr){2-7}

    \grayrow Graph-based & 3.08$\pm$2.38  & 1.85$\pm$1.34  & 2.36$\pm$1.69  & 2.89$\pm$2.44  & 2.62$\pm$2.24  & 2.44$\pm$1.72  \\
ESKF-based & 0.39$\pm$0.20  & 0.35$\pm$0.16  & 0.37$\pm$0.14  & 0.39$\pm$0.15  & 0.43$\pm$0.19  & 0.35$\pm$0.15  \\ 
            \Xhline{3\arrayrulewidth}
        \end{tabular}
    }   
\end{table*}

To visually illustrate the distinction between the two algorithms, we focused on a specific scenario for each dataset: mmMAGIC-Indoor-LOS configuration with five base stations. In Fig.~\ref{fig:2D_figs}, we present visual representations of the trajectories in the xy-plane (upper row) and the z-coordinate over time (lower row) for different datasets. Each subfigure corresponds to a specific dataset (V101, V102, etc.), and within each subfigure, three trajectories are displayed:
\textbf{Blue}: Ground-truth trajectory, \textbf{Red}: Graph-based estimation, \textbf{Green}: ESKF estimation.

\begin{figure*}[!t]
    \centering
    \begin{subfigure}{0.32\textwidth}
        \includegraphics[width=\linewidth]{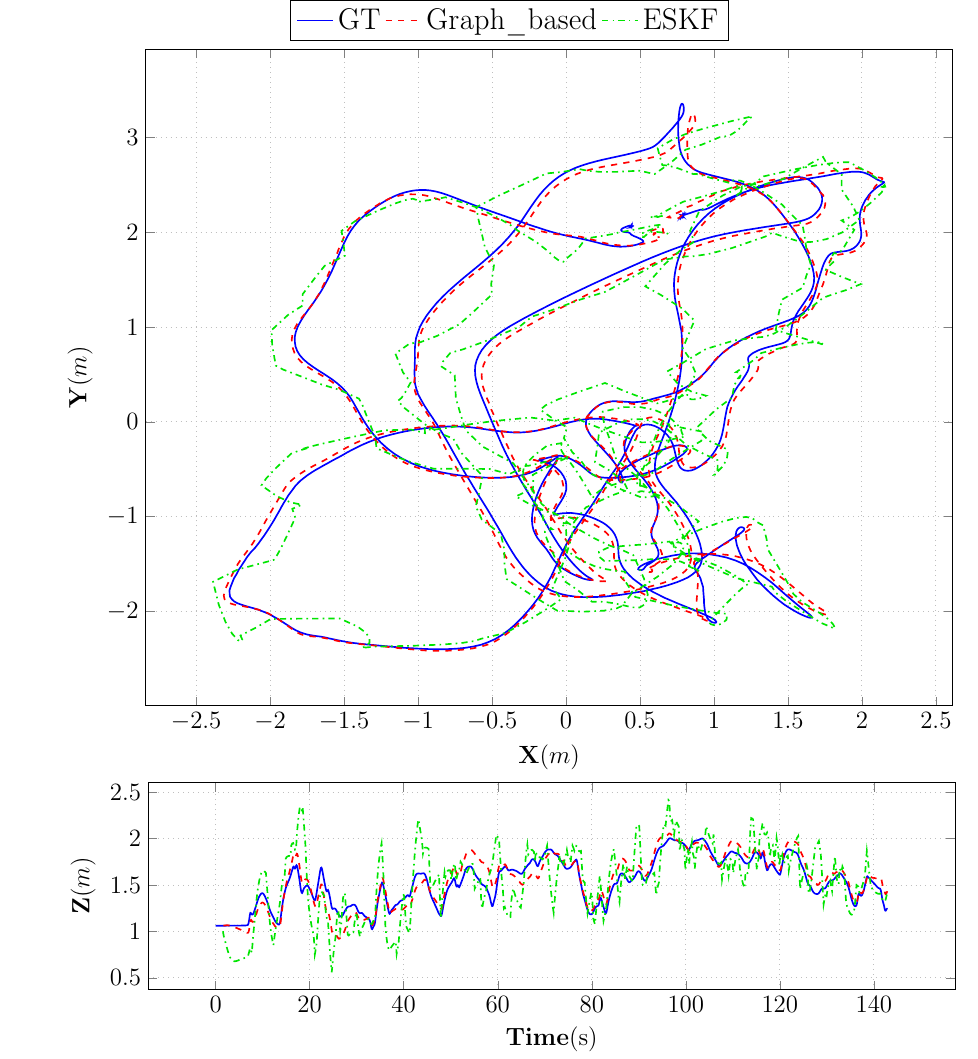}
        \caption{V101}
    \end{subfigure}
    \begin{subfigure}{0.32\textwidth}
        \includegraphics[width=\linewidth]{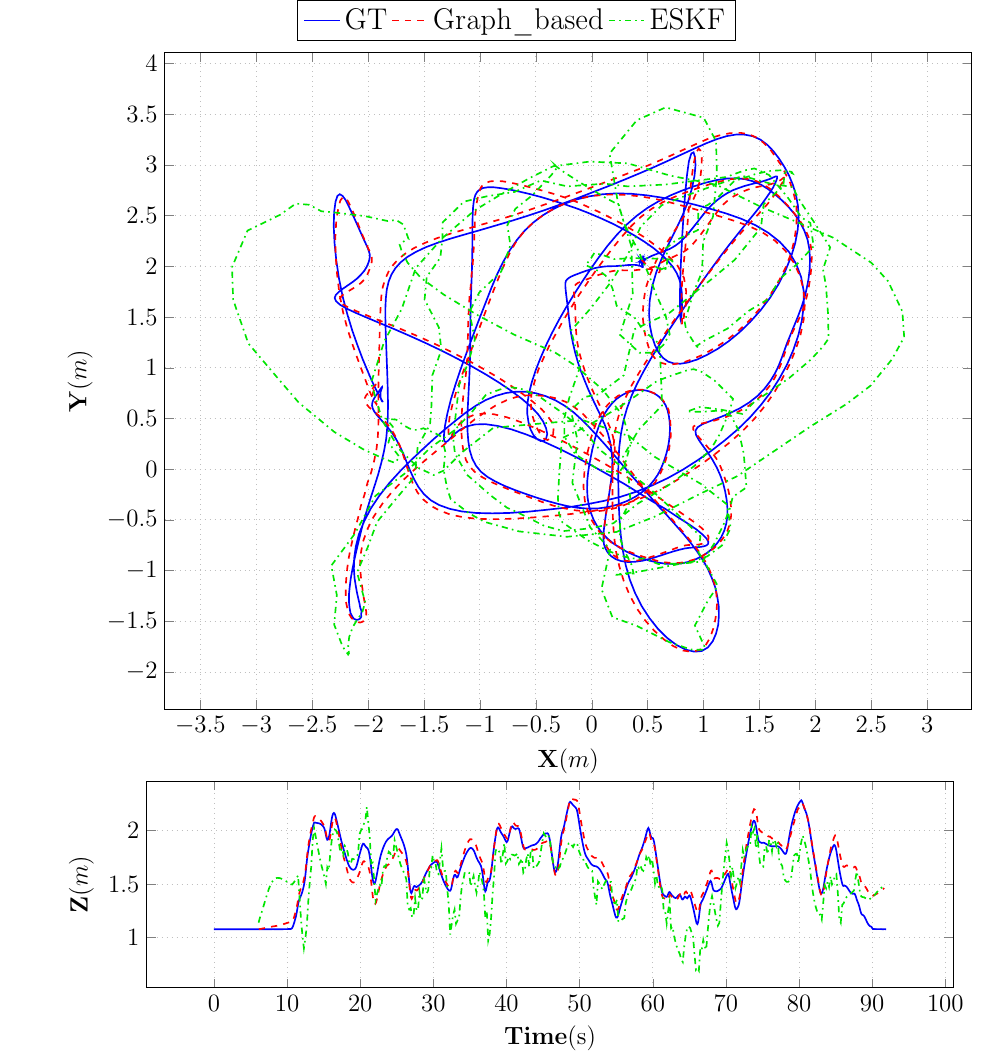}
        \caption{V102}
    \end{subfigure}
    \begin{subfigure}{0.32\textwidth}
        \includegraphics[width=\linewidth]{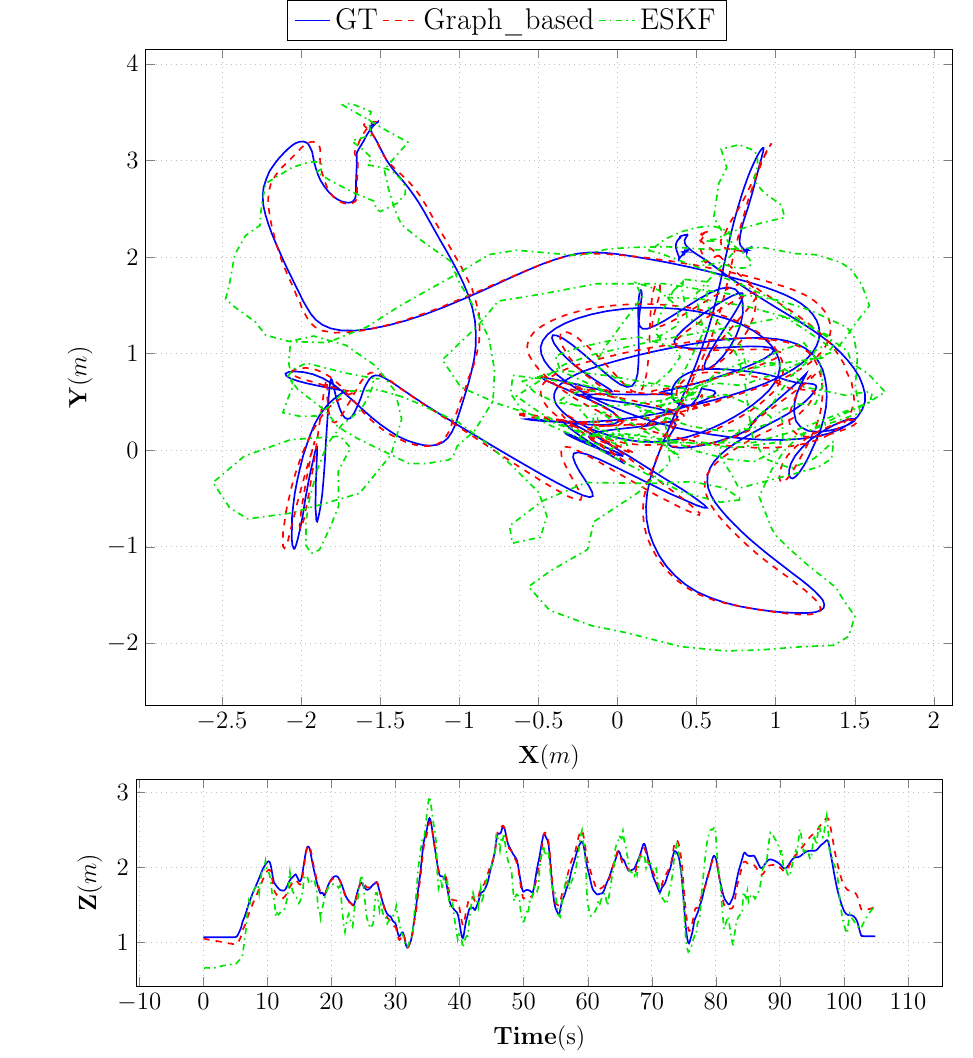}
        \caption{V103}
    \end{subfigure}
    \\
    \begin{subfigure}{0.32\textwidth}
        \includegraphics[width=\linewidth]{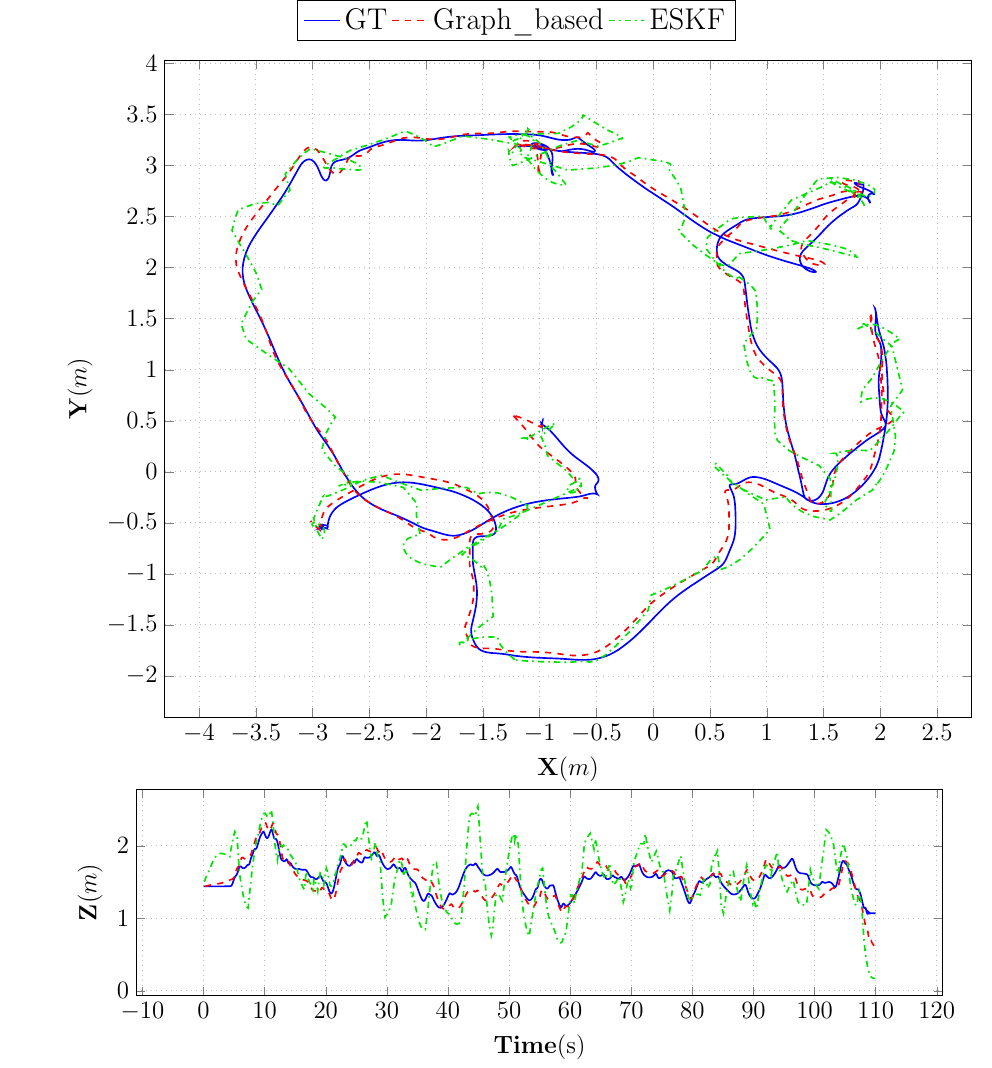}
        \caption{V201}
    \end{subfigure}
    \begin{subfigure}{0.32\textwidth}
        \includegraphics[width=\linewidth]{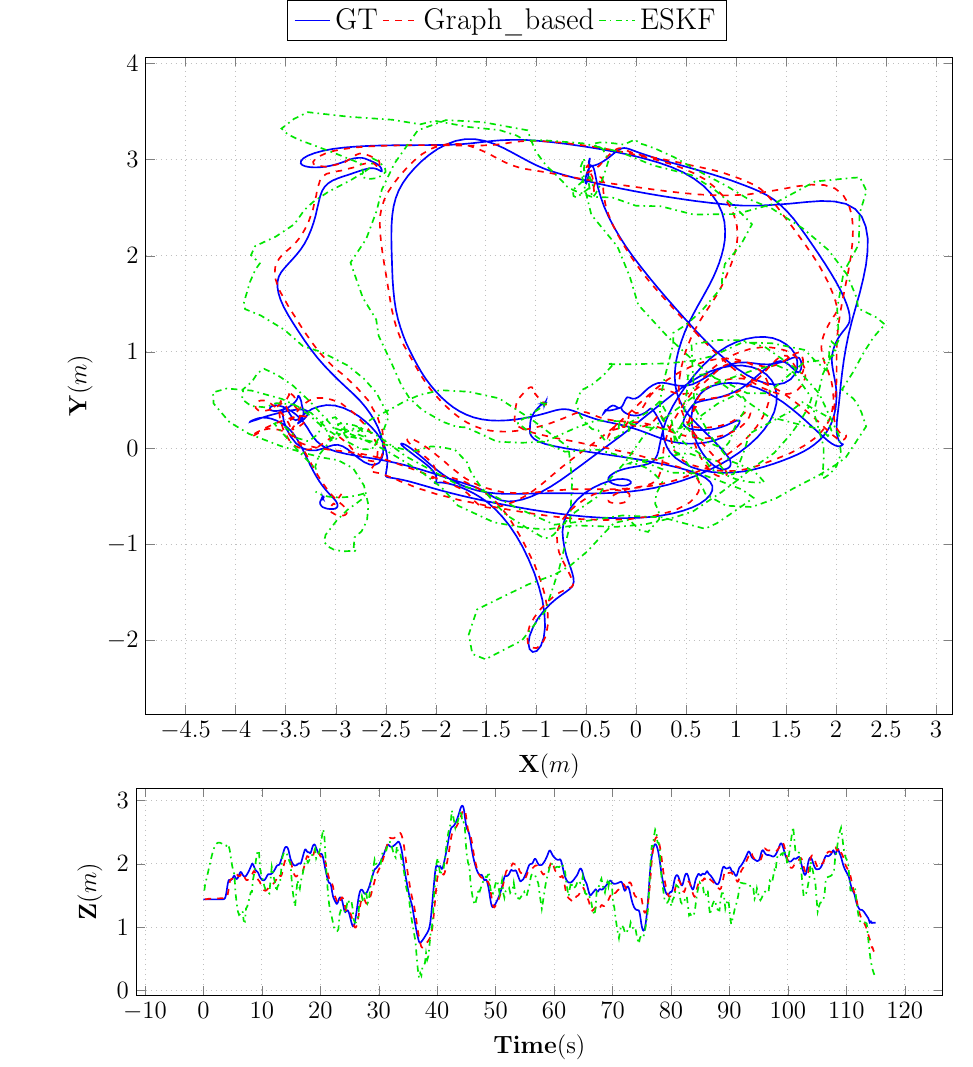}
        \caption{V202}
    \end{subfigure}
    \begin{subfigure}{0.32\textwidth}
        \includegraphics[width=\linewidth]{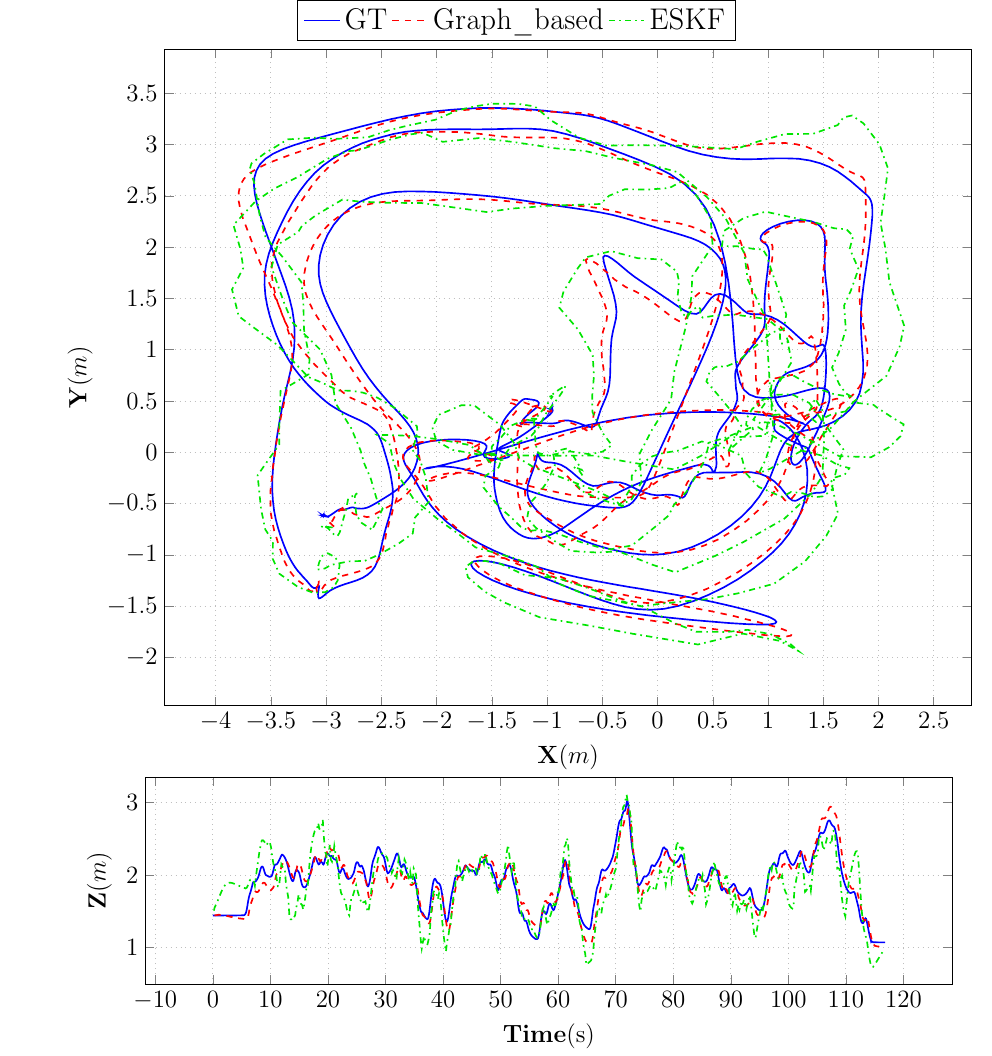}
        \caption{V203}
    \end{subfigure}
    \caption{Trajectories in the xy-plane (upper low) and z-coordinate over time (lower row) for different datasets. Blue represents ground-truth trajectories, red represents graph-based estimation, and green represents ESKF estimation.}
    \label{fig:2D_figs}
\end{figure*}

\subsection{Limitations}

The approach used in this study currently has several limitations and hence leaves potential for future work. The 3D position is not fully constrained with only two antennas, making convergence difficult without other measurements. Nevertheless, the UAV's rotation errors primarily result from IMU noise, as the 5G \toa measurements only provide distances to the antennas. 
The yaw estimation has drift issues because it lacks global measurement to correct it. Integrating other sensors can improve the localization accuracy by observing the rotation around $z$, \eg~employing a magnetometer. Notably, a camera can be incorporated to add other constraints on the 6DoF relative motion based on visual features and loop closures.

Furthermore, the error in the $z$ axis is larger than along $x$ and $y$ axes because of limited offset or variation in the positions of the base stations in the height direction. We foresee possibly fusing the barometer's absolute height measurements to relieve such issues. Additionally, the localization accuracy depends heavily on the quality of the \toa measurements, which can be negatively affected by NLOS conditions. In such cases, correctly setting the measurement uncertainty for each \toa range factor, using Mahalanobis distance to discard outliers, or applying a robust kernel (such as \eg~, a Huber kernel) to the cost function, may be beneficial to alleviate the problem.

Finally, the proposed method assumes that the positions of the base stations are known with high confidence and fixed in the exact location, which may not be the case in real-world scenarios where the stations may be moving or their positions may be completely unknown. However, especially the PGO approach can be adapted to also include an estimation of the unknown position of the BSs, if the \toa measurements can be unambiguously assigned to the sending base station, respectively.

\section{Conclusion and Future Work}
\label{sec:conclusion}
In conclusion, our research demonstrates the potential of utilizing 5G Time of Arrival (ToA) range measurements alongside inertial sensor data for precise indoor localization of Micro Aerial Vehicles (MAVs) across various scenarios and network configurations. We accurately determined MAV positions and orientations by employing graph-based optimization and Error State Kalman Filtering (ESKF). We conducted a thorough comparative analysis across different network scenarios, trajectories, and base station setups.

Our comprehensive investigation revealed valuable insights. We observed that increased base stations and bandwidth improved \toa-based localization accuracy. Despite limitations in the vertical dimension due to base station offsets, \toa measurements exhibited potential for maintaining position estimates at a global level. Graph-based optimization consistently outperformed ESKF, especially in high-bandwidth 5G setups. 

Our research roadmap involves enhancing localization accuracy by incorporating camera visual data. This integration will improve the robustness of our methodology, particularly in scenarios prone to noise or non-line-of-sight conditions. Combining multiple data sources promises to elevate localization accuracy and enable effective drone operation in complex indoor environments.

Our results highlight the potential of graph-based optimization for accurate indoor localization, setting the stage for future research. The synergy between wireless communication and sensor fusion offers breakthroughs in robot pose estimation. In closing, this research emphasizes interdisciplinary collaboration and paves the way for enhancing MAV localization algorithms.

This current approach's limitations invite further research, particularly to elevate localization accuracy. Future work could explore fusing 5G measurements with additional sensors like magnetometers or cameras, potentially mitigating UAV rotation errors and imposing constraints for 6 Degrees of Freedom (6DoF) motion. Additionally, incorporating barometers for absolute height measurements might alleviate z-axis errors arising from limited base station positioning. Strategies like improved uncertainty estimation, outlier rejection, and robust cost functions hold promise for managing noisy Time of Arrival (ToA) measurements, especially in non-line-of-sight scenarios. Finally, adapting localization methods like Pose Graph Optimization to handle unknown or dynamic base station positions could enhance the approach's real-world robustness.

\backmatter







\bmhead{Author contribution}

All authors actively contributed to the study's conception. Meisam Kabiri led the coding, design, and initial draft, with valuable assistance from Claudio Cimarelli. All other authors offered essential input and support, shaping the conceptual framework, providing insights on methodology and the evaluation system, and contributing to manuscript preparation. Additionally, they played a crucial role in editing and reviewing the manuscript. All authors have thoroughly read and given their approval for the final manuscript.



\bmhead{Funding}
This research was funded in whole, or in part, by the Luxembourg National Research 
Fund (FNR), 5G-Sky Project, ref. C19/IS/13713801/5G-Sky. For the purpose of open access, and in fulfilment of the obligations arising from the grant agreement, the authors have applied a Creative Commons Attribution 4.0 International (CC BY 4.0) license to any  Author Accepted Manuscript version arising from this submission.

\bmhead{Availability of data and materials} Not applicable.

\bmhead{Code availability} Not applicable.

\section*{Declarations}

\bmhead{Conflicts of interest} No Conflicts of interest
\bmhead{Ethics approval} Not applicable
\bmhead{Consent to participate} Not applicable
\bmhead{Consent for publication} Not applicable





\bibliographystyle{ieeetr}
\bibliography{Refs}

\end{document}